\RequirePackage{fix-cm}
\documentclass[smallextended]{svjour3}

\smartqed  
\usepackage{graphicx}
\usepackage{cite}

\usepackage{amsfonts}
\usepackage{comment}
\usepackage{amsmath}
\usepackage{color}

\usepackage{enumitem}
\usepackage{hyperref}
\usepackage[normalem]{ulem}

\begin{document}

\title{Multi-target prediction for dummies using two-branch neural networks}

\author{Dimitrios Iliadis*         \and
        Bernard De Baets          \and
        Willem Waegeman 
}

\institute{Dimitrios Iliadis \at
              KERMIT, Department of Data Analysis and Mathematical Modelling, Ghent University, Coupure links 653, B-9000 Ghent, Belgium\\
              \email{dimitrios.iliadis@ugent.be} (*Correspondence)           
           \and
           Bernard De Baets \at
              KERMIT, Department of Data Analysis and Mathematical Modelling, Ghent University, Coupure links 653, \\
              \email{bernard.debaets@UGent.be}           
           \and
           Willem Waegeman \at
              KERMIT, Department of Data Analysis and Mathematical Modelling, Ghent University, Coupure links 653, \\
              \email{willem.waegeman@ugent.be}           
           \and
}

\date{Received: date / Accepted: date}

\maketitle

\begin{abstract}

Multi-target prediction (MTP) serves as an umbrella term for machine learning tasks that concern the simultaneous prediction of multiple target variables. Classical instantiations are multi-label classification, multivariate regression, multi-task learning, dyadic prediction, zero-shot learning, network inference, and matrix completion. Despite the significant similarities, all these domains have evolved separately into distinct research areas over the last two decades. This led to the development of a plethora of highly-engineered methods, and created a substantially-high entrance barrier for machine learning practitioners that are not experts in the field.
In this work we present a generic deep learning methodology that can be used for a wide range of multi-target prediction problems. We introduce a flexible multi-branch neural network architecture, partially configured via a questionnaire that helps end users to select a suitable MTP problem setting for their needs. Experimental results for a wide range of domains illustrate that the proposed methodology manifests a competitive performance compared to methods from specific MTP domains.


\end{abstract}

\section{Introduction}\label{intro}
\raggedbottom
Over the last decade, multi-target prediction (MTP) has emerged as a novel umbrella term, unifying supervised learning techniques that are concerned with predicting multiple target variables at the same time. In principle, these targets can be of different types, such as nominal, ordinal, or real-valued. 
Driven by tutorials and workshops at international conferences, such as ICML 2013 and ECML/PKDD 2014, 2015 and 2018, the area of MTP has attracted significant interest in the machine learning community. Its applicability potential is continuously increasing, as more and more real-world problems require the simultaneous prediction of multiple targets. 

In the field of machine learning one can identify many classical examples of MTP tasks, such as the image tagging task from the area of computer vision \cite{wang2016cnn,wei2015hcp,yan2019multi}, the document tagging task from the field of text mining \cite{chen2017ensemble,huang2019hierarchical}, as well as the product recommendation task that is prevalent in online retailing \cite{fu2018novel,wei2017collaborative}. In addition to these typical examples, one can also identify instances of MTP-related applications that are less well known yet important. In the field of climate science, forecasting the weather in different areas of the world at the same time is a quite complicated task that necessitates the modeling of relationships between various atmospheric processes \cite{papagiannopoulou2018global}. In medicine, patients can usually be associated with multiple interacting pathologies at the same time \cite{baltruschat2019comparison,kumar2018boosted,chen2019deep}. Finally, the emergence of the latest pandemic has highlighted the importance of rapid drug discovery \cite{pliakos2019predicting,rifaioglu2020deepscreen,jin2017multitask}. In this field, the initial goal is to find a set of chemical compounds that show high binding affinity with a biological target, so the use of automated multi-target prediction methods can provide a much-needed speedup.

All these applications are usually encountered in machine learning papers as use cases for specialized techniques. These techniques typically belong to well-known subfields like multi-label classification \cite{yeh2017learning, read2009classifier,tsoumakas2010random,yu2014multi,rokach2014ensemble}, multivariate regression \cite{de2002multivariate,du2017hierarchical,xu2013multi}, multi-task learning \cite{NEURIPS2018_432aca3a,misra2016cross,liu2019end}, dyadic prediction \cite{menon2011link, menon2010log, schafer2015dyad}, hierarchical multi-label classification \cite{wehrmann2018hierarchical, cerri2014hierarchical}, zero-shot learning \cite{romera2015embarrassingly, norouzi2013zero}, matrix completion \cite{jain2013low,shan2010generalized}, and hybrid matrix completion \cite{strub2016hybrid,dong2017hybrid}, which from a distance all look quite different from one another. A recent survey \cite{waegeman2019multi} reviewed not less than 100 methods from these subfields from a general multi-target prediction perspective. In addition, a formal mathematical framework to gather those subfields under a single umbrella was expounded. 

The said mathematical framework will be the point of departure for the goal of the present paper, which is the development of a general deep learning methodology for multi-target prediction problems. Instead of introducing a method that achieves state-of-the-art performance for a narrow range of problems, we present a flexible two-branch neural network architecture that is applicable in a wide range of MTP problems. This type of architecture shows some resemblance with a few deep learning methods that have been recently proposed for specific tasks, such as collaborative filtering \cite{he2017neural, wang2019neural} and metric learning \cite{hoffer2015deep,yi2014deep,mueller2016siamese}. However, we are the first to make this architecture generally accessible for a wide range of multi-target prediction problems. We make the methodology user friendly by introducing a small questionnaire that supports a semi-automated configuration of the two-branch neural network by means of small modifications in its architecture, loss function and inputs. In this way we unlock multi-target prediction to a wide range of users with basic machine learning expertise. 

Our can see some parallels between our work and an ongoing trend in deep learning research towards the development of general-purpose neural network architectures instead of architectures that are only useful for a specific problem setting. For example, the chapter on recurrent and recursive nets in the book of Goodfellow et al.~\cite{goodfellow2016deep} discusses general deep learning architectures for sequence modelling tasks, of which one-to-one, one-to-many, and many-to-many architectures of equal or different length are specific instantiations. Other well-known examples of general-purpose machine learning methodologies are structured support vector machines \cite{wang2009automatic, zhang2011structured}, conditional random fields \cite{lafferty2001conditional,zheng2015conditional} and probabilistic graphical models \cite{frey2005comparison,frey2005comparison}. Especially in statistics it is very common to develop general-purpose frameworks, see e.g.\ generalized linear models \cite{mccullagh2019generalized}. Those models can be applied to various types of supervised learning problems, such as binary and multi-class classification problems, as well as regression problems involving real-valued, ordinal and count-based targets. 

This paper is organized as follows. \hyperref[sec:sec2]{Section 2} quickly reviews the mathematical framework 
of~\cite{waegeman2019multi}, which unifies a wide range of multi-target prediction problems. That section also discusses the inner workings of our proposed questionnaire. \hyperref[sec:sec3]{Section 3} explains several examples of real-world tasks and details how the questionnaire can help with selecting the most suitable MTP problem setting. \hyperref[sec:sec4]{Section 4} presents a detailed view of the two-branch neural network architecture while emphasizing the main characteristics of its flexibility. \hyperref[sec:sec5]{Section 5} gives a summary of closely-related work. \hyperref[sec:sec6]{Section 6} showcases that the proposed methodology works well for a wide range of problems, by comparisons with 15 different methods on 21 different datasets, across 6 MTP problem settings. In the \hyperref[sec:sec7]{last section}, we formulate a conclusion and some future perspectives, discussing the current limitations of our work.

\section{Towards a rule-based system for MTP problem setting selection}
\label{sec:sec2}
In this section we introduce the MTP framework, as well as the novel questionnaire we designed in order to identify the proper problem setting. We also detail the four validation settings that are used in the area of~MTP.

\subsection{The MTP prediction framework}
Let us start with the formal definition of a multi-target prediction problem, as introduced in \cite{waegeman2019multi}. 

\begin{definition}\label{def:MTP_definition}
A multi-target prediction problem is characterized by instances $\mathbf{x} \in \mathcal{X}$ and targets $\mathbf{t} \in \mathcal{T}$ with the following properties:
\begin{itemize}
\itemindent=10pt   
  \item [(P1)] A training dataset $\mathcal{D}$ is comprised of triplets $(\mathbf{x}_i,\mathbf{t}_j,y_{ij})$,
  where $\mathbf{x}_i$ represents an instance ($i\in\{1,\ldots,n\}$), $\mathbf{t}_j$ represents a target ($j\in\{1,\ldots,m\}$), 
  and $y_{ij} \in \mathcal{Y}$ is the score that quantifies their relationship. This dataset can be arranged in an $n \times m$ matrix $\mathbf{Y}$
  that is usually incomplete.
       \item [(P2)] The score set $\mathcal{Y}$ consists of nominal, ordinal or real values.  
    
    \item [(P3)] The objective is to predict the score for any instance-target couple $(\mathbf{x},\mathbf{t}) \in \mathcal{X} \times \mathcal{T}$.
\end{itemize}
\end{definition}

Intentionally, this definition is kept very general in order to cover a wide range of MTP settings. In \cite{waegeman2019multi} also formal definitions are given for the most common settings, grouped into three categories:
\begin{itemize}
\item MTP settings without any kind of usable features (side information) for the targets: this includes the more conventional settings, such as multi-label classification, multivariate regression and multi-task learning.
\item MTP settings with side information for targets: this includes settings such as hierarchical multi-label classification, dyadic prediction, multi-task learning with task features, zero-shot learning and matrix completion with side information. 
\item Non-MTP settings: these are settings that could be expressed as multi-target prediction settings, but are not covered for technical reasons. Two such cases are multi-class classification and structured output prediction.    
\end{itemize}

We do not repeat all those definitions here, but refer the interested reader to Appendix A. However, going over the various definitions is not unimportant in view of understanding the purpose of the questionnaire that is introduced next. So, let us see what we get for by far the most popular setting in literature, namely multi-label classification. 

\begin{definition}
The \textbf{multi-label classification} setting is an instance of the MTP framework with the following additional properties:
\begin{itemize}[leftmargin=*,align=left]
\itemindent=10pt    \item [(P4)] All targets are observed during training ($|\mathcal{T}| = m$).
\item [(P5)] No side information is available for targets, thus we identify them with natural numbers ($\mathbf{t}_j = j$).
 \item [(P6)] The score matrix $\mathbf{Y}$ is fully observed.
 \item [(P7)] The score set is $\mathcal{Y}=\{ 0, 1 \}$.
\end{itemize}
\end{definition}

One can see that for multi-label classification three additional properties appear, in addition to the four general properties that hold for all MTP problems. In Appendix A we provide similar definitions for multivariate regression, multi-task learning, hierarchical multi-label classification, dyadic prediction, zero-shot learning, and matrix completion with and without side information. All those settings have some specific properties, and the purpose of the questionnaire will be to map the answers of users to such properties.

\subsection{The rule-based system}\label{val_setting_sec}

We are able to propose the appropriate MTP problem setting using a rule-based system deployed on-top of a purpose-built questionnaire. The questionnaire is partly answered automatically with our framework from the characteristics of the dataset. There are also questions that currently can only be answered by the user and that have been carefully designed to extract his/her intentions about the given problem. We imagine that by using a graphical interface that accepts the test set, a future version can automatically detect whether the user expects a generalization to unseen instances or targets. In the current stage of development, we use the following questions:

\begin{itemize}
\itemindent=8pt    \item [\textbf{Q1}:] Is it expected to encounter novel instances during testing? \textbf{(yes/no)}
\itemindent=8pt    \item [\textbf{Q2}:] Is it expected to encounter novel targets during testing? \textbf{(yes/no)}
\itemindent=8pt    \item [\textbf{Q3}:] Is there side information available for the instances? \textbf{(yes/no)}
\itemindent=8pt    \item [\textbf{Q4}:] Is there side information available for the targets? \textbf{(yes/no)}
\itemindent=8pt    \item [\textbf{Q5}:] Is the score matrix fully observed? \textbf{(yes/no)}
\itemindent=8pt    \item [\textbf{Q6}:] What is the type of the target variable? \textbf{(binary/nominal/ordinal/real-valued)}
\end{itemize}

\begin{table}[]
\centering
    \caption{}
    \label{table:questionnarie_table}
    \begin{tabular}{c|c|c|c|c|c|c}
    \textbf{Q1} & \textbf{Q2} & \textbf{Q3} & \textbf{Q4} & \textbf{Q5} & \textbf{Q6} & \textbf{MTP method}        \\
    \hline
    yes         & no          & yes         & no          & yes         & binary      & Multi-label classification ~\cite{yan2019multi, hua2020relation, baumel2017multi}\\
    yes         & no          & yes         & no          & yes         & real-valued & Multivariate regression~\cite{raj2020multivariate, qi2016household}    \\
    yes         & no          & yes         & no          & no          & -           & Multi-task learning        \\
    yes         & no          & yes         & yes (hierarchy)   & yes         & binary      & Hierarchical Multi-label classification  ~\cite{wehrmann2018hierarchical, zhang2017hierarchical, chen2019deep} \\
    yes         & no          & yes         & yes         & no          & -           & Dyadic prediction~\cite{jin2017multitask, liu2017computational}       \\
    yes         & yes         & yes         & yes         & no           & -           & Zero-shot learning~\cite{romera2015embarrassingly, mishra2018generative}         \\
    no          & no          & no          & no          & no          & -           & Matrix completion~\cite{zhao2016matrix}        \\
    no         & no         & yes           & yes           & no          & -           & Hybrid Matrix
    completion~\cite{deldjoo2019movie} \\
    yes        & yes         & yes          & yes           & no        &  -           & Cold-start Collaborative filtering~\cite{wei2016hybrid} \\
    yes         & no          & yes         & no          & yes         & nominal/categorical           & Multi-dimensional classification~\cite{jia2020multi, bielza2011multi, shatkay2008multi}
    
    \end{tabular}

\end{table}

These questions are designed to determine the possibility of encountering novel instances or targets during the test phase, the availability of usable side information in the form of relations or representations for instances and targets, the sparsity of the score matrix and the type of values inside the matrix. The aforementioned questions generate 128 different combinations. We have internally annotated the most popular cases with the appropriate multi-target prediction setting (see Table~\ref{table:questionnarie_table}), thus transferring our expert knowledge into the rule-based system. 
There are, however, some specific combinations of characteristics that make the resulting example unable to be annotated. These examples usually try to generalize to novel instances or targets without providing the appropriate side information.

The mentioned differences in the availability of side information that is traditionally associated with each MTP problem setting has led to the distinction of several validation settings. In order to support the different inference cases of all the MTP problem settings, we define the following four experimental settings under which one can make predictions for new couples $(\mathbf{x}_i,\mathbf{t}_j)$:
\begin{itemize}
    \item \textbf{Setting A}:  Both $\mathbf{x}_i$ and $\mathbf{t}_j$ are observed during training.
    \item \textbf{Setting B}:  All targets $\mathbf{t}_j$ are observed during training and the goal is to make predictions for unseen instances $\mathbf{x}_i$.
    \item \textbf{Setting C}:  All instances $\mathbf{x}_i$ are observed during training and the goal is to make predictions for unseen targets $\mathbf{t}_j$.
    \item \textbf{Setting D}:  Neither $\mathbf{x}_i$ nor $\mathbf{t}_j$ is observed during training.
\end{itemize}

\begin{figure}
  \includegraphics[trim=0 20 0 20,clip,width=\linewidth]{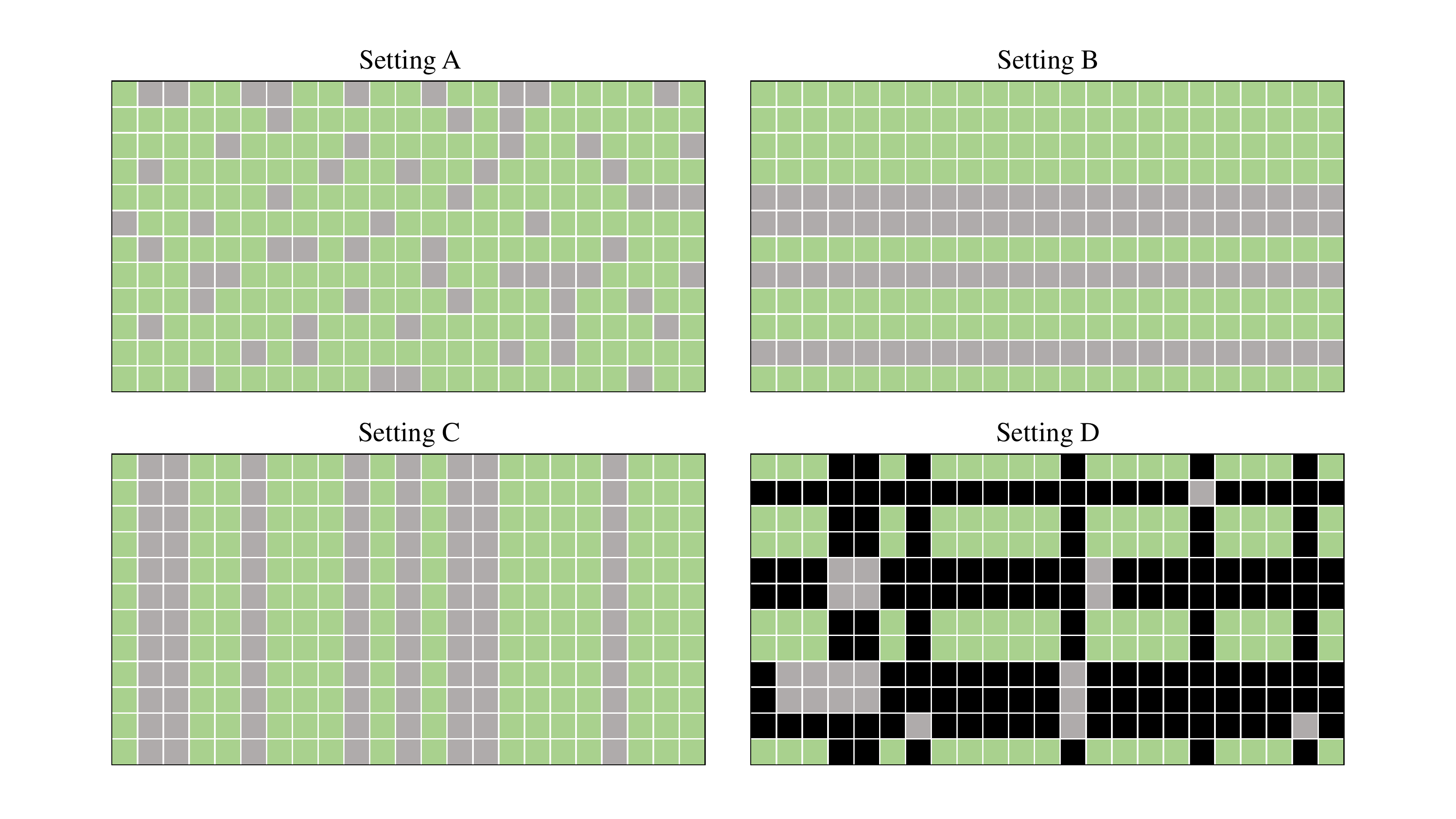}
\caption{The four validation settings supported by the DeepMTP framework visualized for the same interaction matrix. Each row corresponds to a different instance $\mathbf{x}_i$ and each column to a different target $\mathbf{t}_j$. Cells coloured in green correspond to known values $y_{i,j}$ present in the training set. The grey cells represent the missing values or values belonging to the test set. Every black cell in setting D is purposely excluded from both train and test sets. In Setting A the test set is formed by randomly sampling couples $(\mathbf{x}_i,\mathbf{t}_j)$ from the interaction matrix. In Setting B the test set is comprised of entire rows of the interaction matrix, which translates to all possible couples $(\mathbf{x}_i,\mathbf{t}_j)$ for specific targets $\mathbf{t}_j$. Setting C can be seen as the converse of Setting B, as in this case the test set includes all possible couples $(\mathbf{x}_i,\mathbf{t}_j)$  for specific instances. Finally, in Setting D the test set has to contain couples $(\mathbf{x}_i,\mathbf{t}_j)$  of which both instance and target are excluded from the train set.}
\label{fig:5}
\end{figure}

Problems like multi-label classification, multivariate regression, and multi-task learning are mainly associated with Setting B, as they are inductive w.r.t. instances and transductive w.r.t. the targets. This means that during testing, the model is expected to encounter previously-unseen instances, while all targets will be known beforehand. This characteristic informs us about the user's intentions and is determined by two of the questions in our questionnaire, specifically \textbf{Q1} and \textbf{Q2}. But, despite the intentions of the user, his/her answers to questions \textbf{Q3} and \textbf{Q4} are what determines the feasibility of generalization. A basic rule one can use is that if we want to achieve generalization to new instances (targets), appropriate side information should be available for those instances (targets). This is why Setting A is usually associated with matrix completion, as in this problem setting no side information is available for instances or targets and thus no generalization is possible for either of them. Finally, Setting D is considered the most challenging of the settings, as the goal is to make predictions for pairs of unseen instances and targets. In the literature on multi-task and transfer learning, this setting is known as zero-shot learning. 

\section{From real-world problems to MTP problem settings}
\label{sec:sec3}
This section details real-world examples that map to four of the most popular MTP problem settings (multi-label classification, dyadic prediction, matrix completion, and multi-task learning). For each of these examples, we explain how specific characteristics of the datasets and common requests from the end-user provide answers to the queries of our purpose-built questionnaire. Readers already familiar with the various MTP settings might consider to skip this section. 

\subsection{Multi-label classification}
\begin{figure}
  \includegraphics[trim=120 70 120 60,clip,width=\linewidth]{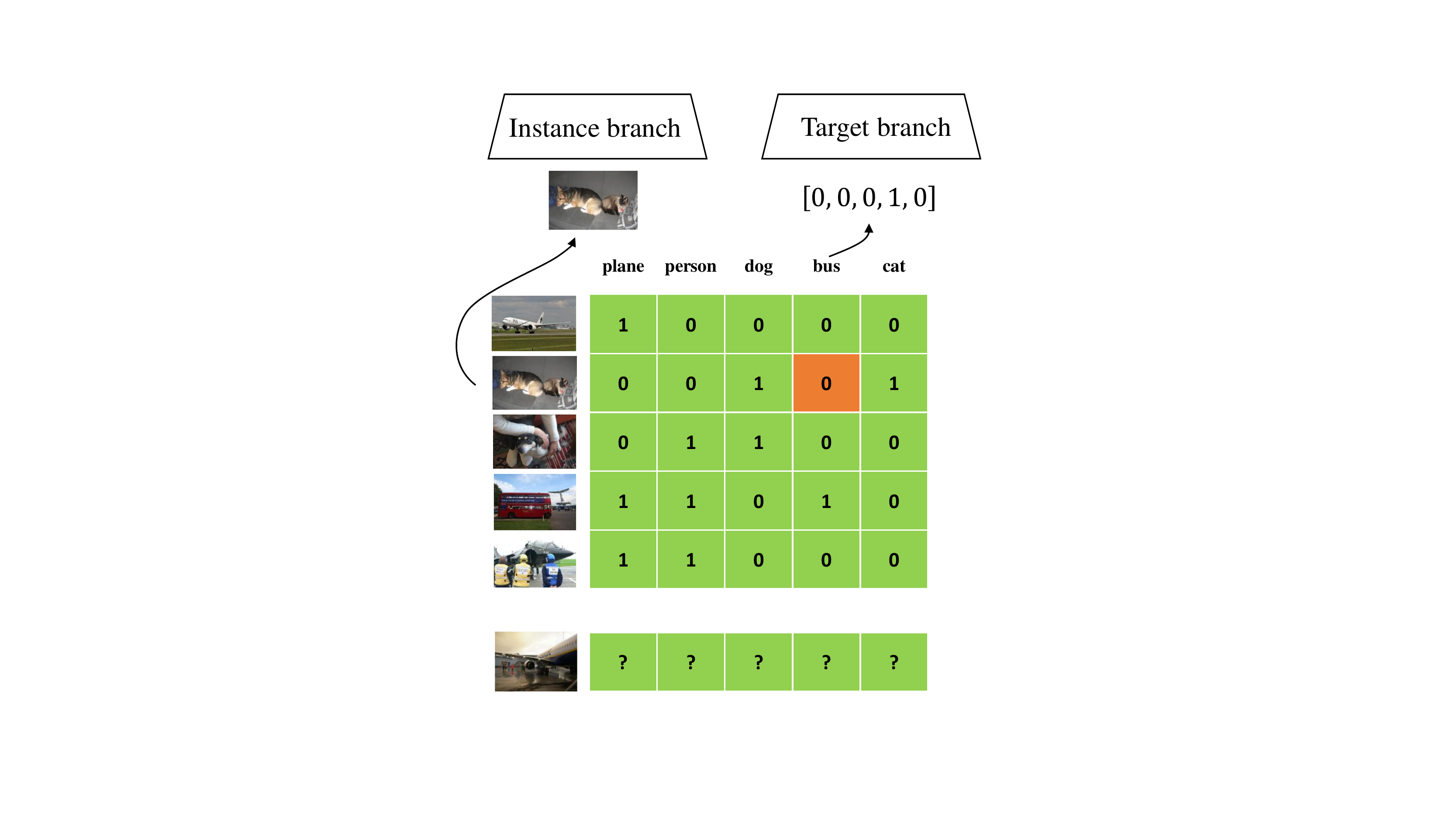}
\caption{Example of a multi-label classification problem from the area of image tagging. Rows represent the different images and columns represent all the possible labels that can be associated with an image. The interaction matrix is fully observed with binary values (1 if the label is associated with the image, 0 otherwise). Side information for the instances corresponds to the raw pixel values of the image. Additionally, because tags  are usually not described by side information, we automatically generate one-hot encoded vectors. After the model is trained, testing involves predicting for a new image whether each of the tags should be associated with it.}
\label{fig:multi-label_classification} 
\end{figure}

A typical example of a multi-label classification problem is that of image tagging shown in Figure~\ref{fig:multi-label_classification}. A user of our framework who wishes to solve a similar problem will have to possess a dataset that contains images (instances) and their known annotations from a set of possible tags (targets). His/her goal will be to annotate new images (\textbf{Q1=yes}) with the tags that were available in the training set (\textbf{Q2=no, Setting~B}).
The pixel values of the images constitute the side information for the instances (\textbf{Q3=yes}) in our DeepMTP framework. At the same time, because the tags usually do not contain any kind of side information (\textbf{Q4=no}), we have to produce one-hot encoded vectors in order to feed the corresponding branch of our neural network. The one-hot encoded vectors have the same length as the total number of targets and all the positions except one are filled with zeros. The position that maps to the unique id of a target is filled with a one. The problem is considered as a classification problem because the tags have a binary relationship with a given image; they can either be associated with that image or not (\textbf{Q6=binary}). The combination of all those characteristics and the specific answers they correspond to in our questionnaire leads us to the identification of the task as a multi-label classification problem. 

It is important to point out that there are also instances of similar image tagging tasks that also offer a tag hierarchy (\textbf{Q4=yes=}\textbf{hierar\-chy}). In such an example, all the other characteristics are the same as what we presented in the paragraph above. Instead of creating a standard one-hot encoded vector, we use the position of each target inside the given hierarchy to create a new vector that is passed to the corresponding branch. The availability  of additional side information for the targets sets this task apart as a hierarchical multi-label classification problem. Information in the form of a hierarchy might also appear in other MTP problem settings such as multivariate regression, but we are not aware of any publicly-available datasets or even research areas with appropriate naming.

\subsection{Dyadic prediction}
Dyadic prediction problems can be found in the field of drug discovery and, more specifically, in the task of predicting the interaction between chemical compounds and proteins (drug-target interaction prediction or DTI). A typical dataset in this area contains interaction information in the form of real-valued affinity scores (\textbf{Q6=real-valued}) between proteins (instances) and chemical compounds (targets). Usually, both of these types of molecules are described by vector representations (\textbf{Q3=yes, Q4=yes}) that can be found in popular databases (PubChem~\cite{kim2021pubchem}, DrugBank~\cite{wishart2006drugbank}, ChEMBL~\cite{gaulton2012chembl}). In a real-world environment, a user, usually a scientist working on a particular disease, identifies a new protein as a potential target for that disease. His/her goal is to check the degree of interaction of that new protein (\textbf{Q1=yes}) with every chemical compound in the aforementioned chemical library (\textbf{Q2=no, Setting B}). The combination of the dataset's properties with the needs of the user leads us to characterize the task as a dyadic prediction problem. It is useful to note that we could easily interchange the role of the proteins and the chemical compounds in our framework, while still considering it a dyadic prediction problem.

\begin{figure}[h]
  \includegraphics[trim=140 40 140 40,clip,width=\linewidth]{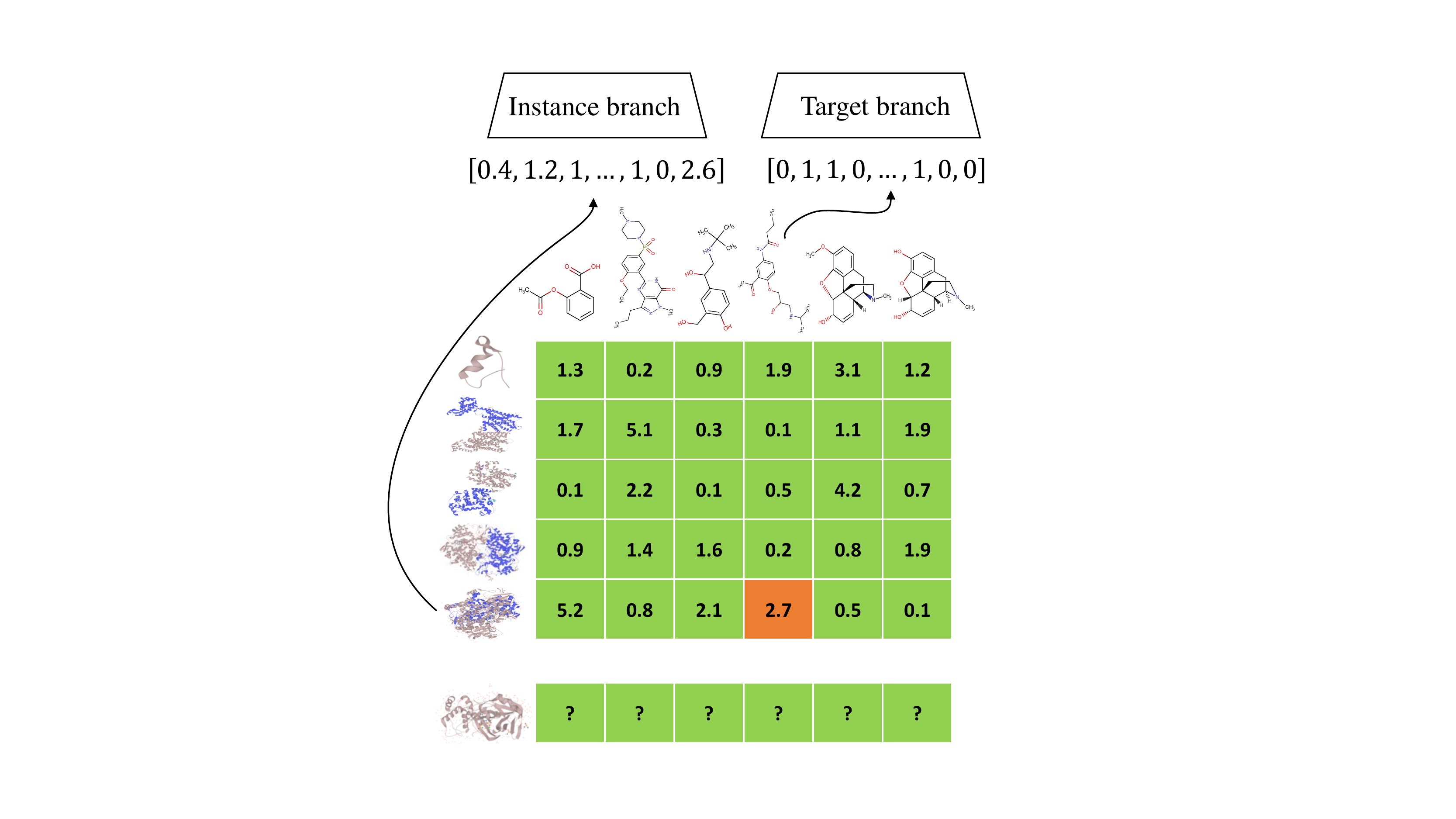}
\caption{Example of a dyadic prediction problem from the field of drug-target interaction prediction. Rows represent the different proteins and columns represent the chemical compound library that a pharmaceutical company might have. The interaction matrix is fully observed and every real value corresponds to the binding affinity of the drug-protein pair.}
\label{fig:dyadic_prediction} 
\end{figure}

\subsection{Matrix completion}
The wide-spread acceptance of e-commerce by companies and customers alike has already generated a significant amount of data that can be used to individualize product recommendations. This has resulted in rapid advancements in the area of recommender systems, which aim to predict the users' interests and recommend items that are likely to be interesting to them. A typical dataset from this area of matrix completion contains some kind of interaction between users (instances) and items (targets). This interaction can be expressed in terms of a binary value (someone bought a product or not) (\textbf{Q6=binary}) or a real value (someone gave a rating to a movie) (\textbf{Q6=ordinal}). 

Another characteristic of this type of dataset is that there is information for only a subset of all the possible pairs (\textbf{Q5=no}). For example, it is only natural that a user cannot rate every movie in a library of thousands. The objective of this task is to make recommendations by completing the interaction matrix that the already seen users (\textbf{Q1=no}) and items (\textbf{Q2=no}) create, while no side information is known for either of them. When side information is available (user's profile page and/or general information about the movie-series) it can be used to potentially improve the performance in the completion task (Hybrid Matrix Completion).

\begin{figure}[h]
  \includegraphics[trim=120 80 120 60,clip,width=\linewidth]{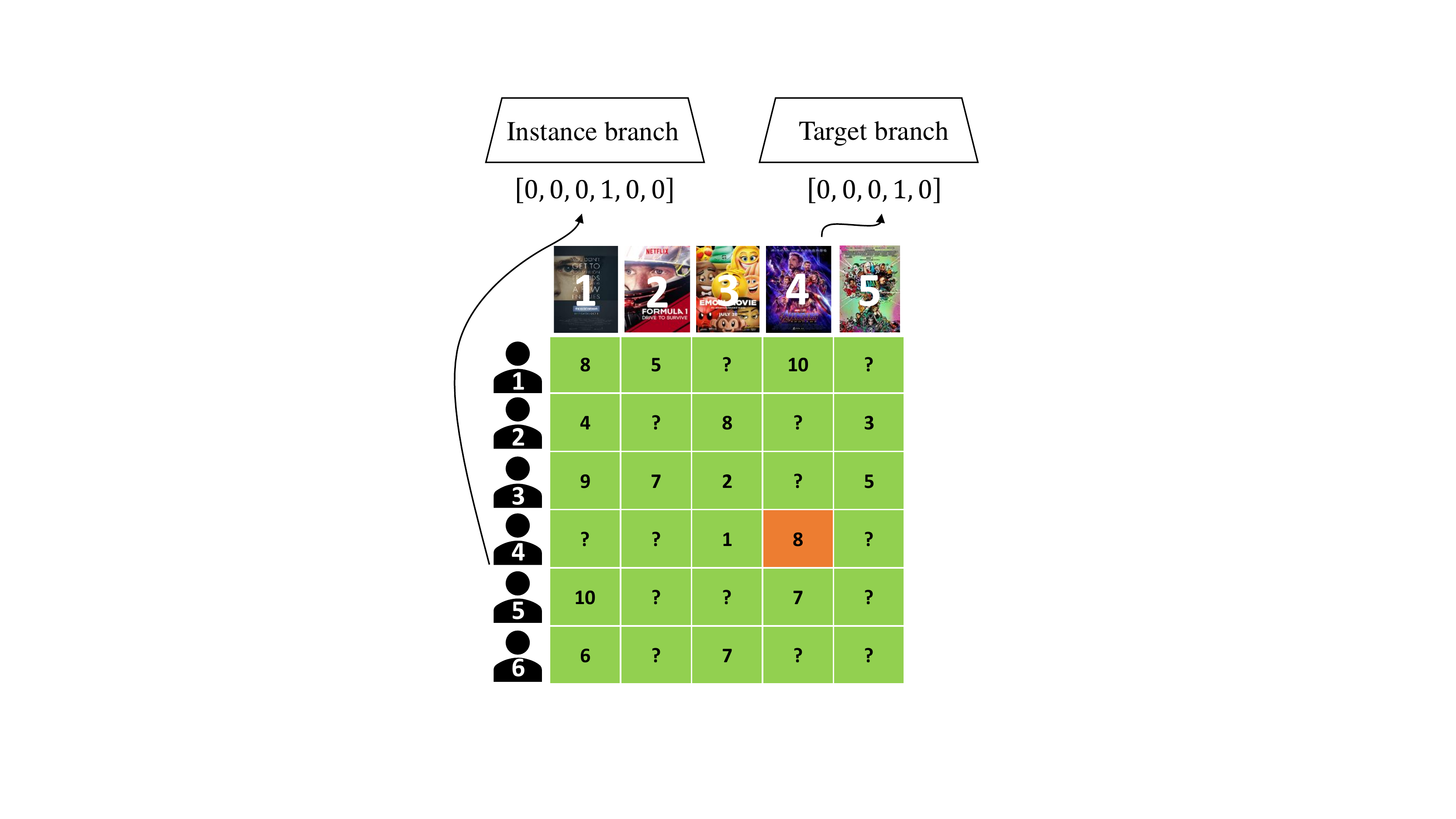}
\caption{Example of a matrix completion problem from the broader area of collaborative filtering. Rows represent the different users of a streaming company and columns represent digital content that belongs to its library. Values inside the interaction matrix represent the ratings that the users have given to the content. In this problem, it is expected that the interaction matrix has mainly missing values as it is impossible for a user to rate every movie and series of the company's library. In the standard matrix completion setting, users and movies are not described by side information, so our framework uses their unique id to construct one-hot encoded vectors. 
The absence of side information also limits the model to only predict ratings for pairs with known users and movies. When side information is actually available, it is possible to extend prediction to pairs with previously unknown users or movies.}
\label{fig:matrix_completion} 
\end{figure}

An extension of this formulation leads to the Cold-start Collaborative filtering problem that can be seen as the result of the continuously-evolving nature of the user-base of many companies. This necessitates the prediction of interactions for new users that were not present in the dataset that the original model was trained on (\textbf{Q1=yes}). By reversing the role of instances and targets, the same argument could be made for new items (\textbf{Q2=yes}) that are added to the database of a company. For example, when a new movie is available on a platform, the objective could be to first predict the expected rating of each user, and then suggest it to the ones that would give high ratings. 
Such a generalization is only possible if the appropriate side information becomes available (\textbf{Q1=yes} and \textbf{Q3=yes}; or \textbf{Q2=yes} and \textbf{Q4=yes}).

\subsection{Multi-task learning}
In contrast to well-defined MTP problem settings like multi-label classification and multivariate regression, multi-task learning contains multiple sub-categories of problems. It thus is more challenging to give a concise definition. A large proportion of work published in this area actually works on problems containing different types of variables for each task (heterogeneous tasks). The pairwise manner in which DeepMTP performs training combined with the use of a single type of loss function during the entire training phase makes the heterogeneous setting incompatible. For example, if our architecture was trained for a multi-task learning problem with two heterogeneous tasks (one binary and one real-valued), we would need two different loss funtions (BCE for the values in the binary task and MSE for the real values in the regression task). This is  currently not possible; in the next section, we will explain that our neural network architecture optimizes only one loss per problem.

\begin{figure}
  \includegraphics[trim=100 50 100 100,clip,width=\linewidth]{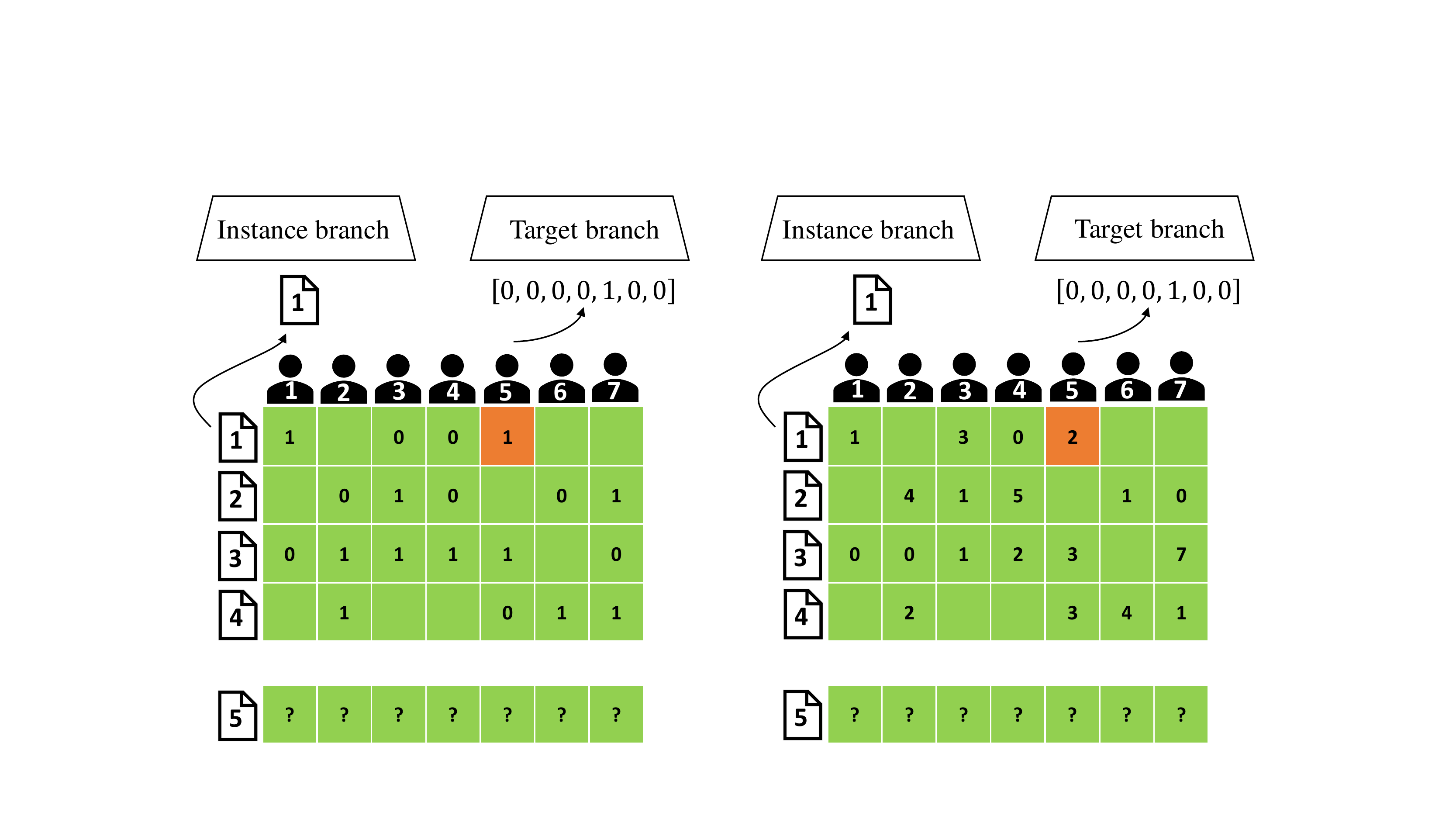}
\caption{Examples of multi-task learning problems from the field of crowdsourced annotation. The figure on the left maps to a binary multi-task learning problem because the values in the interaction matrix are binary. The figure on the right represents a multi-class multi-task learning problem, as the values in the interaction matrix are nominal. All the other characteristics in both figures are identical.}
\label{fig:multitask_learning}       
\end{figure}

A task that suits this setting's characteristics can be found in the area of crowdsourced annotation~\cite{liu2018interactive}.  The quality of training data has been a major limiting factor for the improvement of performance in supervised and semi-supervised tasks. The increasing size of datasets, combined with the high cost of annotating, has led many researchers and companies to crowdsourcing. A user that has a dataset that needs to be annotated can use a crowdsourcing service in order to obtain labels. The resulting dataset he/she will get back could be arranged in an interaction matrix, where the instances map to the original samples of the dataset and the targets map to the annotators. Figure~\ref{fig:multitask_learning} shows a similar example where the instances correspond to documents for which we have the raw text (\textbf{Q3=yes}), and the targets correspond to users that are identified by their id (\textbf{Q4=no}). Depending on the number of possible labels that a user can assign to a document, the interaction matrix can have binary (Figure~\ref{fig:multitask_learning}, left) (\textbf{Q6=binary}) or nominal (Figure~\ref{fig:multitask_learning}, right) 
(\textbf{Q6=nominal}) values. Such a dataset with binary annotations leads to a binary multi-task learning problem, while multi-class annotations lead to a multi-class multi-task learning problem.

A binary multi-task version can also be created if we replace every user's original annotation with a binary value that expresses whether the annotation is correct. Because the size of datasets that need to be annotated is usually close to hundreds of thousands or even millions, it is not feasible for every user to annotate every sample (\textbf{Q5=no}). Finally, during inference, the goal could be to predict how every known user (\textbf{Q2=no}) would annotate a new, previously unseen document or even if these annotations would be correct.

\section{A two-branch neural network architecture for MTP}\label{multi-branch_NN_architecture}
\label{sec:sec4}
The baseline architecture of our framework was first popularized by the neural collaborative filtering (NCF) framework~\cite{he2017neural} in the field of recommender systems. The architecture successfully approximated standard matrix factorization techniques and showed state-of-the-art performance on benchmark datasets. In this work, we show how we can enhance the basic principles of the NCF framework in order to build a generalized framework that achieves a competitive performance in all the settings that fall under the umbrella of~MTP. 

\begin{figure}
  \includegraphics[trim=50 350 50 160,clip,width=\linewidth]{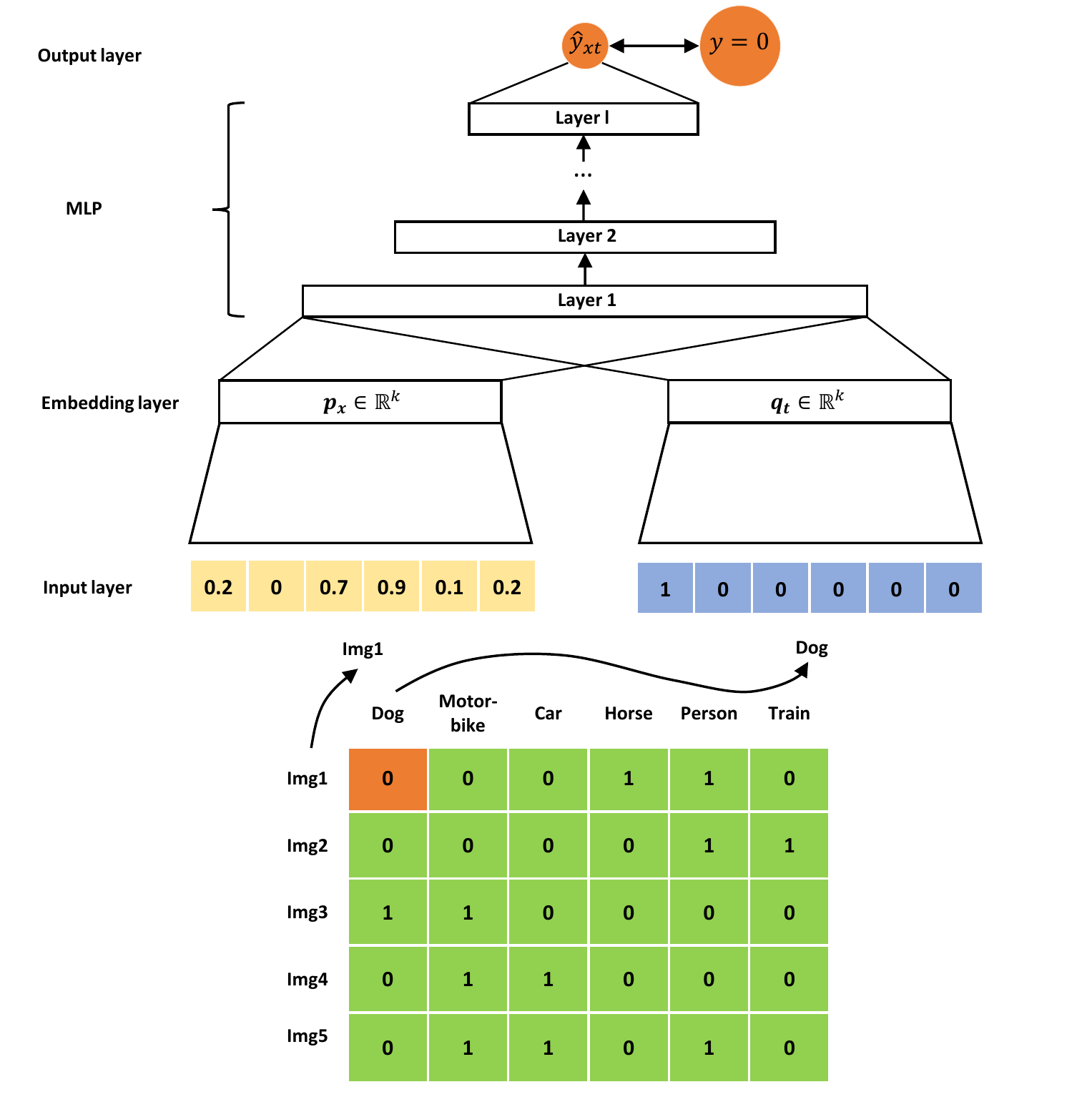}
\caption{Detailed view of the two-branch neural network. The specific example shows an image tagging problem
where the one-hot encoded vectors for the pixel values and the tag are fed into the corresponding branches, 
transformed into embedding vectors, and then passed to an MLP that outputs the predicted interaction score.}
\label{fig:4}
\end{figure}

In the proposed architecture shown in Figure~\ref{fig:4}, the network uses two branches to encode the inputs. More specifically, the bottom input layer of each branch is comprised of two feature vectors $\mathbf{x}_i$ and $\mathbf{t}_j$, which describe the instance and target of a sample in an MTP problem. Both vectors can be customized to support a range of different MTP formulations. For example, in a typical multi-label classification problem, a one-hot encoded vector will be generated to represent a specific target and used as input to the corresponding branch. Using the same principles, in a typical matrix completion problem, we will have to generate one-hot encoded vectors for both instances and targets using their unique ids, very similar to what NCF does.

\begin{figure}
  \includegraphics[trim=0 130 0 130,clip,width=\linewidth]{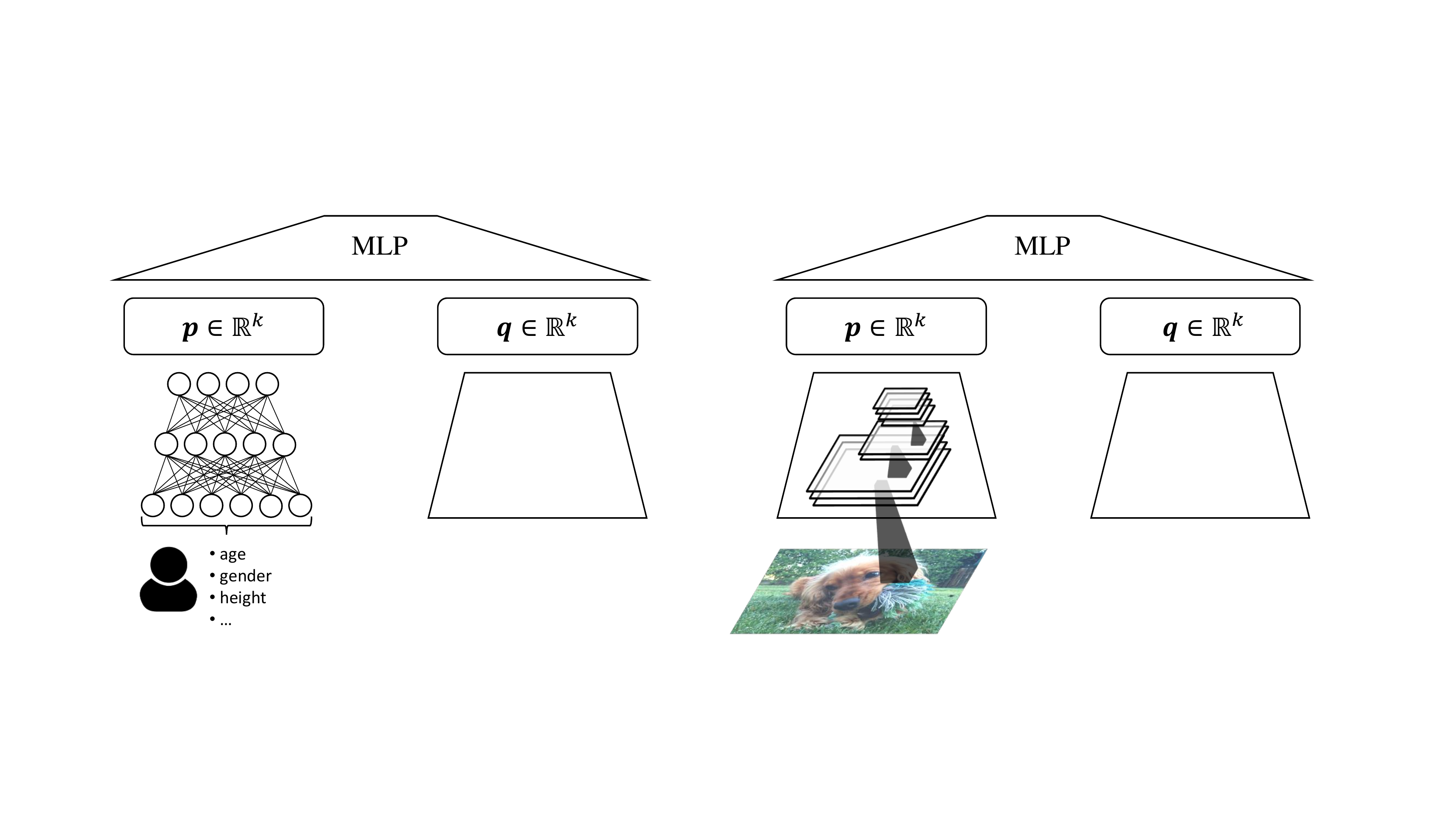}
\caption{Examples of different architectures that can be used in the branches of the multi-branch neural network. In the left figure,
we use a conventional fully connected neural network because the input consists of tabular user-related features,
whereas in the right figure, we use a convolutional architecture because the input is in the form of images. Both versions of our dual branch architecture utilize a final multi-layer perceptron (MLP) that takes as input the vector obtained by concatenating the instance embedding vector $\mathbf{p_x}$ and the target embedding vector $\mathbf{q_t}$}.
\label{fig:3}       
\end{figure}

Above the input layer, we extend the NCF framework by using different types of layers or even entire sub-architectures to better encode the different kinds of inputs the framework may encounter. In cases where no side information is provided (for example, the labels in a multi-label classification problem), we use a single fully-connected layer to project the sparse one-hot encoded input vector to a dense embedding. Otherwise, when explicit side information is available, we have multiple options, depending on the type of input, from several fully-connected layers (tabular health record data, Figure~\ref{fig:3} left) to more specialized architectures based on convolutional neural networks (Figure~\ref{fig:3} right) or graph neural networks (hierarchies). The goal of the embedding layer in both cases is to project the instances and targets to a lower-dimensional latent space, similarly to what is done with the users and items in the product recommendation problem in NCF~\cite{he2017neural}.

The instance embedding $\mathbf{p_x}$ and target embedding $\mathbf{q_t}$ are then concatenated and passed through a multi-layer neural network architecture that maps the embeddings to the predicted target value in the following way:  
\begin{equation} \label{eq1}
\begin{split}
\mathbf{z}_1 & = \phi_1(\mathbf{p_x}, \mathbf{q_t}) = \left[\begin{array}{c}\mathbf{p_x} \\\mathbf{q_t} \end{array}\right]\,, \\
\phi_2(\mathbf{z}_1) & = \alpha_{2}(\mathbf{W}_{2}^T \mathbf{z}_1 + \mathbf{b}_2)\,, \\
& ......\\
\phi_L(\mathbf{z}_{L-1}) & = \alpha_{L}(\mathbf{W}_{L}^T \mathbf{z}_{L-1} + \mathbf{b}_L)\,, \\
\hat{y}_\mathbf{xt} & = \sigma(\mathbf{h}^T \phi_L (\mathbf{z}_{L-1}))\,,\\
\end{split}
\end{equation}
\noindent where $\mathbf{W}$, $\mathbf{b}$ and $\alpha$ correspond to the weight matrix, bias vector and activation function of the final multi-layer perceptron (MLP) layer. We mainly use the leaky rectified linear unit (Leaky ReLU) as activation function in our framework, but because we also perform experiments with custom architectures from third parties instead of the branches, other activation functions may also be utilized (for example, standard ReLU in Resnet~\cite{he2016deep}).

This MLP architecture is able to model more complex, non-linear instance-target relationships compared to a simpler dot product. Even though this idea was popularized by the NCF framework and widely adopted by the CF community, there has been recent work proposing that the dot product may be highly competitive and cheaper to train~\cite{rendle2020neural, dacrema2021troubling}. Regardless, we decided that all the experiments shown below should use an MLP and that we will investigate whether the dot product can be a viable alternative 
for the MTP settings in future work. 

The final output layer consists of a single node that outputs the predicted score $\hat{y}_{\mathbf{xt}}$. In the classification-related MTP settings a sigmoid function is used before the output in order to restrict it to $[0,1]$.
We facilitate training using different loss functions to accommodate the different categories of MTP problem settings. 
In classification problems, training is achieved using the binary cross-entropy loss function:
\begin{equation}
    \uppercase{{L}}_{{\mathrm{BCE}}} =
    -{ \sum_{({\mathbf{x}},{\mathbf{t}}, y) \in \mathcal{D}} {y} \log{\hat{y}_{\mathbf{xt}}} + (1 - y)  \log{(1 - \hat{y}_{\mathbf{xt}}})}\,.
\end{equation}

On the other hand, in problems that fall into the regression category, we use the squared error loss:
\begin{equation}
\uppercase{{L}}_{{\mathrm{MSE}}} = \sum_{({\mathbf{x}},{\mathbf{t}}, y) \in \mathcal{D}} {(y - \hat{y}_{\mathbf{xt}})^2} \,.
\end{equation}
\noindent In both loss functions, $\mathcal{D}$ denotes the set of known interactions in the training set.

In order to make it more accessible to the reader
how training and inference work in our architecture, we make a comparison with a standard neural network in the popular multi-label classification case shown in Figure~\ref{fig:4}. The basic neural network will have as many input nodes as instance features and as many output nodes as there are labels (six in the example). This means that for the example in Figure~\ref{fig:4}, the neural network will use the pixel values of an image as  input and then output the prediction for every label simultaneously. This procedure is followed during training as well as inference. In our architecture, training and inference are performed in a pairwise manner. Instead of working with all the labels of an image simultaneously, we process each instance-target pair separately. Thus, for the same example we detailed earlier, our network will have to input the same image six times to the instance branch and modify the one-hot encoded vector that is passed to the target branch.

\begin{figure}
  \includegraphics[trim=0 150 0 130,clip,width=\linewidth]{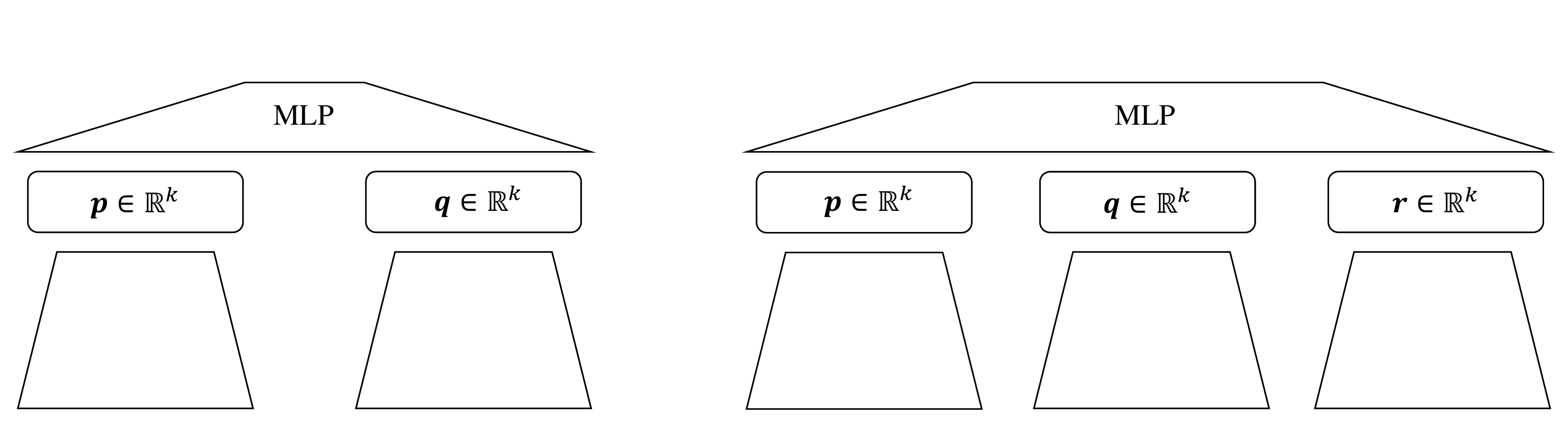}
\caption{General two-branch architecture (left) and tri-branch architecture (right).}
\label{fig:2}       
\end{figure}

It is also important to point out that there are cases in which additional side information is available. These features are usually available for every couple $(\mathbf{x}_i,\mathbf{t}_j)$ in the dataset and have been coined dyadic features in the literature~\cite{van2017mistar}. 
Such information requires an extension of our two-branch architecture by a third branch that allows to encode those dyadic features (Figure~\ref{fig:2} right). Similar architectures have been successfully deployed in tensor factorization applications~\cite{wu2018neural, schreiber2020avocado}. In this setting, training and inference remain largerly unchanged, the only difference being the concatenation of three embedding vectors $\mathbf{p_x}$, $\mathbf{q_t}$ and $\mathbf{r_d}$ instead of just two.

Finally, our neural network architecture, combined with the pairwise manner in which we train our models, allows to make predictions for all four validation settings shown in Figure~\ref{fig:5} (Settings A, B, C and D) without having to make modifications in the core training and inference steps. The only stages in the pipeline that need to be adapted are the preparation of the dataset splitting as well as the computation of the performance metrics. In the experiments presented in Section~\ref{results}, we only report results for Settings A and B, as they are the most frequently encountered ones. In future work, we intend to also report the performance for the two other settings and discuss the differences between them.

\section{Related work}\label{results2}
\label{sec:sec5}
This section's goal is to discuss related work. The literature on multi-target prediction is vast, so we will focus on deep learning approaches for multi-target prediction. First, we review two-branch neural network architectures that have been introduced for specific problem settings (some of thse settings fall under the MTP umbrella). Those architectures are often very similar to the architecture we propose. Second, we review other deep learning methods that can be used for multi-target prediction, i.e.\ architectures that are not based on two branches. Third, we briefly discuss some well-known MTP methods that are not at all based on neural networks. 

Two-branch neural network architectures have been developed for distance metric learning, similarity learning and object matching problems. In these application domains, such architectures are often referred to as Siamese neural networks~\cite{bromley1993signature}. The architecture typically consists of two identical branches, which are both capable of learning the hidden representation of an input vector. The two outputs are then compared, usually through cosine similarity, and the output of such a network can be thought of as the semantic similarity between the two embedding vectors. Siamese neural networks have found extensive use in video analysis~\cite{ryoo2018extreme,liu2017provid}, but also in audio processing~\cite{pitt2005buckeye,chen2011extracting} and natural language processing~\cite{yih2011learning,marelli2014sick,das2016together}. For a more extensive review of the application of Siamese networks, we refer to~\cite{chicco2021siamese}.

Two-branch neural networks can also be used to learn the similarity between two objects of a different type. In this setting, the two branches will have a different architecture, similar to our framework. In computer vision one can find several papers that adopt such an idea for different applications, without a focus on developing general-purpose tools. Convolutions are used in the branch that encodes images, while other layer types are considered in the second branch. As representative examples, let us discuss three papers a bit more in detail.  Wang et al.~\cite{wang2018learning} investigate two-branch neural networks to learn the similarity between image and text modalities for the purpose of phrase localization and bi-directional image-sentence retrieval. 
Shao and Qian~\cite{shao2019three} consider a two-branch convolutional neural network to classify facial expressions. The first branch takes as input the raw image and extracts global features, while the second uses local binary pattern features to extract local texture features. As a third example, 
Pan et al.~\cite{pan2018learning} introduce DualCNN for various low-level vision problems like super-resolution, noise/artifact removal, image deraining, and dehazing. Their architecture consists of two branches, one shallow sub-network to estimate the structures of the input image and one deep sub-network to estimate the details. 

In recent years two-branch neural networks have also been introduced in recommender systems. In fact, the neural collaborative filtering framework of~\cite{he2017neural}, which has been explained in Section~4, has become one of the most popular neural-network-based matrix factorization methods. One of the methods proposed in \cite{he2017neural}, called generalized matrix factorization, computes the dot product between the two branches, but this is only possible when the learned embeddings of the two branches have the same dimensionality. Moreover, the dot-product is not parameterized by any additional (learnable) parameters, which might hamper the predictive performance. That is why they also suggest a modification that is used in our work in which the learned embeddings are concatenated to a single vector that serves as input for another fully-connected feed-forward neural network. As another alternative, He et al.~\cite{he2018outer} use an outer product to explicitly model the pairwise correlations between the dimensions of the embedding space. The outer product creates a two-dimensional interaction map that is then processed by a convolutional neural network to learn high-order correlations among the embedding dimensions effectively.

A natural extension of the use of two branches for matrix factorization is the inclusion of a third branch that can encode a third dimension and thus be used for tensor factorization. Wu et al.~\cite{wu2019neural} introduce a neural-network-based tensor factorization model that contains a third (LSTM-based) branch to characterize the multi-dimensional temporal interactions for relational data. For some applications, we believe that it is also  relevant to include a third branch in our MTP framework, and this is something we will experiment with in the future.

In multi-label classification, several deep learning methods that do not consider two branches have been presented. Gong et al.~\cite{gong2013deep} used a convolutional architecture, similar to what we did for any task that involved images and experimented using different ranking-based loss functions. He demonstrated that the weighted approximated ranking loss, which specifically optimizes the top-$\mathcal{k}$ accuracy (not possible with the current version of our work), works well for multi-label annotation problems. Nam et al.~\cite{nam2017maximizing} propose a sequence-to-sequence recurrent neural network as an alternative to the well-known classifier chains method. Similar to other chaining methods, this neural network is mainly useful for optimizing the subset zero-one loss (not considered in this paper). 
In the area of multi-label image classification, Wang et al.~\cite{wang2016cnn} combine deep convolutional and recurrent neural networks in a framework that is able to learn a joint image-label embedding that exploits label dependencies. Because this approach uses LSTMs, a predefined label ordering is required during training, something that is usually not available. For that reason, Chen et al.~\cite{chen2018order} investigate the effectiveness of a deep learning model that combines a visual attention model with an LSTM and thus does not require any predetermined label ordering. Huynh and Elhamifar~\cite{huynh2020shared} consider a shared multi-attention mechanism that predicts all seen and unseen labels in an image, something that other attention-based approaches are unable to do. Finally, custom architectures have also been proposed for cases in which the number of labels becomes very large (extreme multi-label classification). Liu et al.~\cite{liu2017deep} used deep convolutional neural networks for multi-label text classification and showed a competitive performance for datasets with up to 670k labels. In the same area, Zhang et al.~\cite{zhang2018deep} established an explicit label graph to better model the label space of extreme multi-label classification datasets. Our approach is able to scale linearly with the number of labels, but further work will be needed to improve speed and make experimentation with larger datasets feasible.

Lastly, as far as software packages go, we could not find any work that provides methods for more than two MTP problem settings. Tsoumakas et al. developed Mulan~\cite{tsoumakas2011mulan}, an open-source Java library that implements several transformation methods like binary relevance, label powerset, and other multi-label algorithms like multi-label $k$ nearest neighbors, random $k$-labelsets, the hierarchy of multi-label learners algorithm, and back-propagation multi-label learning. In contrast to the command-line interface of Mulan, MEKA~\cite{read2016meka} is another popular Java library that provides a graphical interface and inherits methods implemented in Weka~\cite{hall2009weka}. Another open-source library that was introduced more recently and is written in python is called scikit-multilearn~\cite{2017arXiv170201460S}. This library can utilize methods from scikit-learn and provides an interface for MEKA, but the set of methods included is limited. Finally, the MLC toolbox~\cite{kimura2017mlc} offered multi-label classification methods for MATLAB/OCTAVE users.

\section{Experimental results on various MTP problems}\label{results}
\label{sec:sec6}
This section's main goal is to convey that our architecture is flexible enough to train and make predictions with minimal configuration changes for multiple MTP problem settings. We also want to showcase that our approach is quite competitive 
with methods that are usually purpose-built for only one of the problem settings. Of course, for the same reason, we do not expect and is generally not our goal to outperform all the methods we are comparing with. At the end of this section, we anticipate that our framework will constitute a viable benchmark for future methods that will be developed for any of the MTP problems settings we have explored. The experimental setup as well as the hyperparameter space of the methods we compare with are located in the \hyperref[hyperparameters_appendix]{Appendix}

\subsection{Multi-label classification}
For the multi-label classifcation problem setting, we selected methods that are available in the scikit-learn~\cite{pedregosa2011scikit} and scikit-multilearn~\cite{2017arXiv170201460S} libraries. More specifically, we compare with a standard neural network in which the number of output nodes is equal to the number of targets, two instances of a binary relevance approach that has a support vector machine (SVM) and a neural network as base classifier, a nearest neighbors method adapted for multi-label classification (MLkNN)~\cite{zhang2007ml}, a multi-output decision tree classifier (DT), and an ensemble of classifier chains (ECC)~\cite{read2009classifier}  that uses an SVM as the base classifier. Experiments were performed using four benchmark datasets from Mulan’s GitHub 
repository~\cite{tsoumakas2011mulan}.  Table~\ref{table:mlc_results_table} lists these datasets along with their main statistics. Because there is no target side information in this setting, our framework uses one-hot encoded vectors as inputs for the corresponding branch.

\begin{table}[t!]
\centering
    \caption{The four multi-label classification data sets used in this study and reported Hamming loss of every method for these datasets.}
    \label{table:mlc_results_table}

\begin{tabular}{lcccc} 
                             & \textbf{Yeast} & \textbf{Scene} & \textbf{Bibtex} & \textbf{Corel5k} \\ \hline
                             &                &                &                 &                  \\
\textbf{data properties}     &                &                &                 &                  \\ \hline
\textbf{\#instances}         & 2417           & 2407           & 7395            & 5000             \\
\textbf{\#instance features} & 103            & 294            & 1836            & 499              \\
\textbf{\#targets}           & 14             & 6              & 159             & 374              \\
\textbf{\#target features}   & 14             & 6              & 159             & 374              \\ \hline
                             &                &                &                 &                  \\
\textbf{Hamming loss}        &                &                &                 &                  \\ \hline
\textbf{BR (SVM)}            & \textbf{0.1935} & \textbf{0.0733}         & \textbf{0.0130} & 0.0146           \\
\textbf{BR (MLP)}            & 0.2223         & 0.0993         & 0.0142          & 0.0201           \\
\textbf{MLP}                 & 0.2406         & 0.0897         & 0.0198          & 0.0183           \\
\textbf{MLkNN}               & 0.202          & 0.0885         & 0.0146           & 0.0162           \\
\textbf{DT}                  & 0.261          & 0.1301         & 0.0137          & 0.0151           \\
\textbf{ECC (SVM)}           & 0.2841         & 0.1312         & 0.0148          & 0.0152           \\
\textbf{DeepMTP}             & 0.2309         & 0.0839         & 0.0157          & \textbf{0.0114}           \\ \hline
\end{tabular}
\end{table}

The hyperparameters of the methods we compare with were optimized through a grid search. The performance metric of choice for this problem setting is the widely used Hamming loss, which the majority of methods can explicitly optimize for. The importance of this characteristic was originally explored in~\cite{dembczynski2010regret} and influenced the selection of the methods we compare with. The results shown in Table~\ref{table:mlc_results_table} illustrate the competitiveness of the DeepMTP framework, as it achieves comparable performance to the other baselines on all four datasets.The experimental section for this MTP problem setting is not as extensive as other papers that are exclusively focused on this area, both in terms of datasets and methods. This is done purposely, as we have to consider many other MTP settings. For extensive comparisons in this area, we refer to work by Madjarov et al.\cite{madjarov2012extensive } and Tsoumakas et al.\cite{tsoumakas2007multi}.

\subsection{Multivariate regression}
Similarly to the previous setting, all the selected methods were obtained from the scikit-learn library~\cite{pedregosa2011scikit}. The methods selected in this setting include a typical multilayer perceptron (MLP), popular single-target approaches (support vector regression (SVR), Kernel ridge regression (KRR), decision tree regression), as well as an ensemble of 50 regressor chain models that use a support vector regressor as base model. We also selected the seven datasets listed in Table~\ref{table:mvr_results_table} from a repository that accompanied~\cite{melki2017multi}.

The hyperparameters of the methods we compare with were optimized through a grid search. The performance metric used in this setting is the commonly-used 
average relative root mean square error (aRRMSE). The results shown in 
Table~\ref{table:mvr_results_table} indicate that our approach is quite competitive, outperforming the other methods on three out of the six available datasets. The DeepMTP framework's performance closely resembles that of the standard neural network and becomes more competitive when the number of training samples increases.

\begin{table}[t!]
\centering
    \caption{The seven multivariate regression data sets used in this study and reported aRRMSE of every method for these datasets.}
    \label{table:mvr_results_table}
\begin{tabular}{lccccccc} 
                             & \textbf{Enb} & \textbf{Jura} & \textbf{Water quality} & \textbf{Oes97} & \textbf{Oes10} & \textbf{Puma8nh} & \textbf{Puma32h} \\ \hline
                             &              &               &                        &                &                &                  &                  \\
\textbf{data properties}     &              &               &                        &                &                &                  &                  \\ \hline
\textbf{\#instances}         & 768          & 359           & 1060                   & 323            & 403            & 8192             & 8192             \\
\textbf{\#instance features} & 8            & 11            & 16                     & 263            & 298            & 8                & 32               \\
\textbf{\#targets}           & 2            & 7             & 14                     & 16             & 16             & 3                & 6                \\
\textbf{\#target features}   & 2            & 7             & 14                     & 16             & 16             & 3                & 6                \\
                             &              &               &                        &                &                &                  &                  \\
\textbf{aRRMSE}              &              &               &                        &                &                &                  &                  \\ \hline
\textbf{SVR/target}          & 0.1161       & \textbf{0.5747}        & 0.9493               & 0.5394            & 0.3410              & 0.8818             & 0.9634           \\
\textbf{DTR}                 & 0.1629       & 0.7587                 & 1.0310               & 0.7620            & 0.5024              & 0.9590             & 1.0949            \\
\textbf{KRR/target}          & 0.1606       & 0.5785                 & 0.9500               & 0.5312            & 0.3468              & 0.8625             & 1.0001              \\
\textbf{MLP}        & \textbf{0.0933}       & 0.6334                 & 0.9809               & 0.7885            & 0.3946              & 0.8750             & 1.0008            \\
\textbf{ECC (SVR)}           & 0.1231       & 0.5751                 & 0.9472               & 0.5393            & \textbf{0.3409}     & 0.8780             & \textbf{0.9633}           \\
\textbf{DeepMTP}             & 0.0954       & 0.6614                 & \textbf{0.9279}      & \textbf{0.4843}   & 0.4292              & \textbf{0.8509}   & 1.002  \\ \hline        
\end{tabular}
\end{table}

\subsection{Hierarchical multi-label classification}
For the hierarchical multi-label classification problem
problem setting, we selected two image classification datasets that included hierarchical information for the targets, which in this case corresponds to tags that can be associated with an image. More specifically, these datasets are the MSCOCO and the VOC 2007, two really popular benchmarks in the area of multi-label classification. Microsoft COCO~\cite{lin2014microsoft} is a benchmark that contains 82081 images in the training set and 40504 images in the validation set. There are also 80 different labels that can be associated with an image with 
the actual average being 2.9 labels per image. 

\begin{figure}
  \includegraphics[trim=50 530 190 0,clip,width=\linewidth]{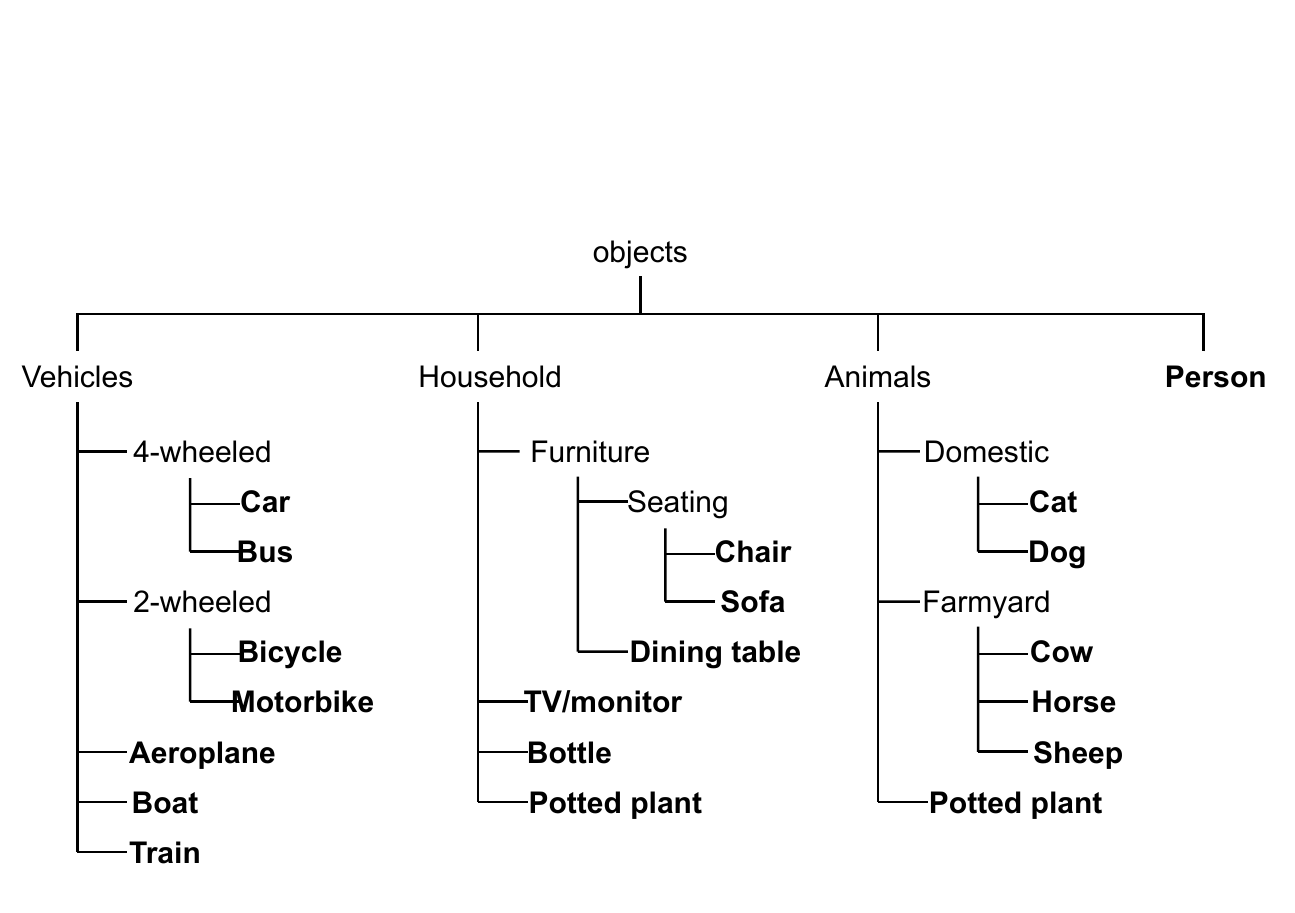}
\caption{Hierarchy of the 20 categories present in the VOC 2007 dataset.}
\label{fig:voc_hierarchy}       
\end{figure}

The second dataset used in this setting comes from the PASCAL Visual Object Classes Challenge (VOC 2007)~\cite{everingham2010pascal} and is divided into train, validation and test sets. This benchmark contains 9963 images and 20 different tags that are organized in the hierarchy shown in~Figure~\ref{fig:voc_hierarchy}.  In our experiments, the methods were trained on both the train and validation sets and the evaluation was done using the test set.

In terms of the configuration of our framework for this problem, we decided to use a pre-trained version of the ResNet-101 architecture, similar to what is shown in the right of Figure~\ref{fig:3}, as well as Figure~\ref{fig:multi-label_classification}. For the branch that encodes the targets, we experimented with two different versions. In the first one, we create standard one-hot encoded vectors, similarly to what we do when no side information is available. In the second version, we utilize the available tag relations by constructing sparse vectors that encode the given hierarchy. For example, inspecting the hierarchy for the VOC 2007 dataset in Figure~\ref{fig:voc_hierarchy}, we count nine categories and 20 final classes-tags. To construct a vector that encodes the hierarchy, we first create a 29-dimensional vector populated by zeros. Each position of the vector maps to a different category or tag. Then, to represent a specific tag we start from the root of the hierarchy and traverse it until we arrive at that tag. For each node we encounter, we assign a one to the corresponding position in the vector.

In terms of methods, we decided to compare with a graph-convolution-based approach that was proposed in~\cite{chen2019multi}. In the latter paper, the authors present experiments with the two datasets presented above. Even if theoretically the use of the same train-test split would not necessitate re-running their experiments in order to compare with our framework, we decided to do so using their published implementation. The results on MS-COCO and VOC2007 are shown in Table~\ref{table:hmlc_mscoco_voc2007_results_table}. In terms of metrics, we decided to use the same as in~\cite{chen2019multi}. More specifically, we computed the macro-wise (macro-P, macro-R, macro-F1) and instance-wise (inst-P, inst-R, inst-F1) versions of recall, precision and F1 score. 

\begin{table}[t!]
\centering
    \caption{Comparison of the ML-GNC and DeepMTP methods in terms of multiple performance metrics on the MS-COCO dataset.}
    \label{table:hmlc_mscoco_voc2007_results_table}
\begin{tabular}{lcc} 
                   & \textbf{MS-COCO} & \textbf{VOC2007} \\ \hline
                   &         &         \\
\textbf{instance-precision} &         &         \\ \hline
\textbf{ML-GNC}             & \textbf{0.858}   & \textbf{0.8725}  \\
\textbf{DeepMTP}            & 0.8344  & 0.8612  \\
                   &         &         \\
\textbf{instance-recall}    &         &         \\ \hline
\textbf{ML-GNC}         & \textbf{0.754}   & \textbf{0.9107}  \\
\textbf{DeepMTP}            & 0.6559  & 0.8035  \\
                   &         &         \\
\textbf{instance-F1}        &         &         \\ \hline
\textbf{ML-GNC}             & \textbf{0.803}   & \textbf{0.8912}  \\
\textbf{DeepMTP}            & 0.7345  & 0.8313  \\
                   &         &         \\
\textbf{macro-precision}    &         &         \\ \hline
\textbf{ML-GNC}             & \textbf{0.851}  & \textbf{0.8602}  \\
\textbf{DeepMTP}            & 0.7885  & 0.8364  \\
                   &         &         \\
\textbf{macro-recall}       &         &         \\ \hline
\textbf{ML-GNC}             & \textbf{0.72}    & \textbf{0.8947}  \\
\textbf{DeepMTP}            & 0.6144  & 0.8004  \\
                   &         &         \\
\textbf{macro-F1}           &         &         \\ \hline
\textbf{ML-GNC}             & \textbf{0.78}    & \textbf{0.8771}  \\
\textbf{DeepMTP}            & 0.6807  & 0.8137 \\ \hline
\end{tabular}
\end{table}

From the results presented above, we observe that the DeepMTP framework shows a competitive performance with the ML-GNC method. It is also important to mention that in our experiments ML-GNC achieved a slightly worse performance across all six metrics compared to what is reported in~\cite{chen2019multi}. The same paper makes comparisons with multiple other methods, for which we could not find any implementation. We hypothesize that this is also the reason why some performance values are missing from their table of results (the source papers report a subset of the six metrics they use, so they could not copy these results). This is the reason why we do not include these methods in the present work. Finally, we report that our experiments with the two different versions of target features did not result in a significant difference in performance. This can be explained by the fact that both of the used hierarchies are quite shallow, and thus do not offer useful information compared to the standard one-hot encoded features. We believe that the ability of our framework to easily include or disregard the hierarchical information boosts the potential of our framework for the hierarchical classification setting.

\subsection{Matrix completion}
For this task, we decided to compare with methods that are available in Microsoft's repository of recommender systems~\cite{microsoft_recommenders}. Because 
we do not support a ranking loss at this stage, we only included methods that optimize for regression. These methods include a matrix factorization approach by alternating least squares (MF-ALS), a neural network approach that is really similar to ours but uses a dot product to combine the instance and target embedding vectors (fastai), the Riemannian Low-rank Matrix Completion (rlrmc), and another neural network approach that trains a wide linear model as well as a deep neural network (wide \& deep). In terms of datasets, we decided to use two versions of the widely-used movielens dataset, one with 100 ratings (movielens100k) and one with one million ratings (movielens1M). The dataset contains ratings that users gave to movies, exactly as shown in Figure~\ref{fig:matrix_completion}. The test set is formed by randomly selecting 25\% of the known ratings (Setting~A).
Because we do not have any side information available for either users (instances) or movies (targets), we generate one-hot encoded vectors for both of them. Similarly to what we see with all the other MTP problem settings, the DeepMTP framework is quite competitive. For both versions of the movielens dataset, the performance is quite similar to or even outperforms that of the other methods.

\begin{table}[t!]
\centering
    \caption{Comparison of the collaborative filtering approaches in terms of multiple regression performance metrics for movielens100k and movielens1M}
    \label{table:mc_movielens100k_results_table}
\begin{tabular}{lcc} 
                      & \textbf{Movielens100k} & \textbf{Movielens1M} \\ \hline
\textbf{}             &                        &                      \\
\textbf{RMSE}         &                        &                      \\ \hline
\textbf{MF-ALS}       & 0.9628                 & \textbf{0.8593}               \\
\textbf{fastai}       & 0.9431                 & 0.8745               \\
\textbf{rlrmc}        & 1.0441                 & 0.8640               \\
\textbf{wide \& deep} & 0.9491                 & 0.9574               \\
\textbf{DeepMTP}      & \textbf{0.9391}                 & 0.8782               \\
\textbf{}             &                        &                      \\
\textbf{MAE}          &                        &                      \\ \hline 
\textbf{MF-ALS}       & 0.7488                 & 0.6783               \\
\textbf{fastai}       & 0.7443                 & 0.6954               \\
\textbf{rlrmc}        & 0.7961                 & \textbf{0.6751}               \\
\textbf{wide \& deep} & 0.7561                 & 0.7771               \\
\textbf{DeepMTP}      & \textbf{0.7421}                 & 0.6904               \\
\textbf{}             &                        &                      \\
\textbf{R2}           &                        &                      \\ \hline
\textbf{MF-ALS}       & 0.2590                 & \textbf{0.4133}               \\
\textbf{fastai}       & 0.2853                 & 0.3869               \\
\textbf{rlrmc}        & 0.1240                 & 0.4016               \\
\textbf{wide \& deep} & 0.2762                 & 0.2651               \\
\textbf{DeepMTP}      & \textbf{0.2913}                 & 0.3817               \\ \hline
\end{tabular}
\end{table}

\subsection{Multi-task learning}

For the multi-task learning problem setting, we decided to experiment with crowdsourcing datasets, in a very similar setup as described in Figure~\ref{fig:multitask_learning}. This was partly done because contemporary research in multi-task learning works on heterogeneous interaction matrices, something that our framework does not support at this moment.

The datasets we used in this setting were first introduced in~\cite{liu2018interactive}. These include two image datasets that are labeled by users. The first one contains 800 images of dogs and 52 annotators that have to label each image with one of four available breeds. The second dataset contains 2000 images of 10 different types of birds that were labeled by 65 annotators. In both datasets, the majority of possible image-user pairs is missing, as it is challenging for a user to annotate thousands of images. To simplify the problem, we used the correct annotations that were supplied for every image to transform the original multiclass, multi-task learning problem into a binary one. A given cell in the final interaction matrix shows whether or not the user labeled an image correctly (Figure~\ref{fig:multitask_learning} left).

In terms of methods we decided to compare with, we chose two baselines. The first one simply predicts the majority class. For example, if the majority of a user's annotations is correct, we predict that he/she will also label all the test set images correctly. The second approach is the standard single-task approach in which we train a single model for every task separately. Because the side information of the instances corresponds to raw images, we chose the VGG architecture instead of the corresponding branch in our two-branch neural network. More specifically, we used a pre-trained version of the VGG-11 architecture that had every layer's weights, except for the last one, freezed. This was intentionally done to improve running time and also because both datasets did not have enough instances to train such a massive architecture.

\begin{table}[t!]
\centering
    \caption{Reported accuracy, AUROC, AUPR of every method on the two multi-task datasets.}
    \label{table:multitask_results_table}
\begin{tabular}{lcc}
                              & \textbf{dogs} & \textbf{birds} \\ \hline
                              &                        &                      \\
\textbf{data properties}      &                        &                      \\ \hline
\textbf{\#instances}          & 800                    & 2000                 \\ 
\textbf{\#instance features}  & 3*224*224              & 3*224*224            \\
\textbf{\#targets}            & 52                     & 65                   \\
\textbf{\#target features}    & 52                     & 65                   \\
                              &                        &                      \\
\textbf{macro-accuracy}       &                        &                      \\ \hline
\textbf{majority voting}      & \textbf{0.7461}                 & 0.6809               \\
\textbf{single-target Resnet} & 0.7043                 & 0.6588               \\
\textbf{DeepMTP}              & 0.6621                 & \textbf{0.7027}               \\
\textbf{}                     &                        &                      \\
\textbf{AUROC}                &                        &                      \\ \hline
\textbf{majority voting}      & 0.5                    & 0.5                  \\
\textbf{single-target Resnet} & 0.5123                 & 0.5105               \\
\textbf{DeepMTP}              & \textbf{0.611}                  & \textbf{0.6637}               \\
\textbf{}                     &                        &                      \\
\textbf{AUPR}                 &                        &                      \\ \hline
\textbf{majority voting}      & 0.7406                 & 0.6192               \\
\textbf{single-target Resnet} & 0.7421                 & 0.6585               \\
\textbf{DeepMTP}              & \textbf{0.8071}                 & \textbf{0.7555}              \\ \hline
\end{tabular}
\end{table}

In terms of results, it is clear that the datasets used are not large enough to train the neural networks. In terms of accuracy, the majority voting approach is competitive as the test sets in most cases were comprised of only a few samples. The single-target Resnet approach was unable to properly train and was only predicting the majority class. In terms of AUROC and AUPR, and for both datasets, our approach clearly outperforms the other two methods.

\subsection{Dyadic prediction}
For the dyadic prediction problem setting, we chose to compare with a network inference approach that uses an ensemble of bi-clustering trees (eBICT)~\cite{pliakos2019network} on datasets that are used in that paper. Although the implementation of the eBICT method is not available online, it was kindly provided to us by the authors upon simple request. The four datasets (see Table~\ref{table:dyadic_results_table}) that we include in our work are heterogeneous interaction networks that are publicly available and commonly used in the field of bioinformatics. For each dataset, the interaction matrix is populated by binary values and side information is available for both instances and targets. Two of the datasets correspond to drug-protein interaction networks and were originally introduced, together with two additional datasets, as a gold standard in the area of DTI prediction. Side information for the drugs amounts to vectors that code for the similarity of their chemical structure, while side information for the proteins comes in terms of similarities based on the alignments of their sequences. The original four datasets were differentiated by the category of the target protein they include: nuclear receptors (NR), G-protein-coupled receptors (GR), ion channels (IC), and enzymes (E). In this work, we excluded two of the datasets (NR and GR) because of their very small size, both in terms of number of instances as well as in terms of number of targets.

The remaining two datasets that we used in our work corresponds to regulatory networks for two different micro-organisms. The first dataset concerns an 
{\em E.~coli} regulatory network (ERN) that contains pairs of transcription factors and genes of the {\em E.~coli} bacterium. The second dataset 
representes a similar network but with genes from the {\em Saccharomyces cerevisiae} yeast. Here, the side information for both instances and targets consists of expression values.

\begin{table}[t!]
\centering
    \caption{Reported micro-AUROC and micro-AUPR of every method on the four dyadic prediction datasets.}
    \label{table:dyadic_results_table}
\begin{tabular}{ccccc} 
                             & \textbf{DPI-E} & \textbf{DPI-IC} & \textbf{SRN} & \textbf{ERN} \\ \hline
                             &                &                 &              &              \\
\textbf{data properties}     &                &                 &              &              \\ \hline
\textbf{\#instances}         & 664            & 204             & 1821         & 1164         \\
\textbf{\#instance features} & 664            & 204             & 9884         & 445          \\
\textbf{\#targets}           & 445            & 210             & 113          & 154          \\
\textbf{\#target features}   & 445            & 210             & 1685         & 445          \\
                             &                &                 &              &              \\
\textbf{micro-AUROC}         &                &                 &              &              \\ \hline
\textbf{eBICT}               & 0.8053         & \textbf{0.8338}          & \textbf{0.8169}       & 0.8536       \\
\textbf{DeepMTP}             & \textbf{0.8571}         & 0.8312          & 0.8166       & \textbf{0.8874}       \\
\textbf{}                    &                &                 &              &              \\
\textbf{micro-AUPR}          &                &                 &              &              \\ \hline
\textbf{eBICT}               & \textbf{0.555}          & 0.4248          & \textbf{0.1686}       & \textbf{0.4450}       \\
\textbf{DeepMTP}             & 0.45           & \textbf{0.4289}          & 0.162        & 0.4295       \\ \hline
\end{tabular}
\end{table}

In terms of performance metrics, we follow what was proposed in~\cite{pliakos2019network}. These include the micro-average versions of the area under the precision recall curve (AUPR), as well as the area under the receiver operating characteristic curve (AUROC). Concerning the hyperparameters for the eBICT method, we used the defaults that were proposed in the corresponding paper. The results shown in Table~\ref{table:dyadic_results_table} demonstrate the competitiveness of our approach. In terms of AUROC, we outperform the eBICT method on two out of the four datasets and show a similar performance on the remaining two. In terms of AUPR, we manage to outperform the eBICT method on only one dataset, but we remain competitive on the other three. At this point, it is important to state that we only report results for only one of the four validation settings (Setting B) even though in~\cite{pliakos2019network} the authors also experiment with Settings C and D. We argue that this is similar to what can happen in real-world situations, as a user can choose only one type of generalization despite more options being available. In future work, we expect to also compare our performance for the other three settings (A, C, and D), similar to what the eBICT paper presents.

\section{Conclusions \& Future Work}
\label{sec:sec7}
In this paper, we proposed a new framework that aims to make all the problem settings that fall under the umbrella of MTP more accessible to the end-user. In order to do so, we introduced a novel, purpose-built questionnaire that distils our understanding of the commonalities and differences that the MTP problem settings display. We also showed examples of how the characteristics of specific real-world problem settings and datasets lead to specific combinations of answers to our questionnaire and ultimately to a specific MTP problem setting being identified.

We then explained how we use the selected MTP setting information to configure a flexible multi-branch neural network. We also showcased all the different modifications we can perform in the network's architecture, starting from how we handle different types of input data to what losses we can use depending on the different types of output values each problem setting offers. Finally, we provided extensive experimental results for five popular MTP problem settings, covering 21 different datasets and 19 different methods. From those results, we are able to show that our architecture can be quite competitive in all five MTP problem settings with minimal modifications, while facing different types of input and output data, different dimensionalities of input features, and different validation strategies. To conclude, we believe that this architecture can be used as a reliable benchmark in future work related to all MTP problem settings.

In terms of limitations, we would like to  point out some variations of MTP problem settings that are not recommended for our framework. Our multi-task experiments included datasets with binary values for all the included targets (binary multi-task learning). Datasets with multiple classes (multi-class multi-task learning) could be tackled by replacing the single output node with a number of nodes that is equal to the number of classes. MTP problem settings like multi-dimensional classification could be tackled using the same configuration for the output layer. Also, similarly to the work by Jia et al.\cite{jia2020multi}, comparisons can be made using modified multivariate regression datasets and baseline methods that are used in multi-label classification (Binary Relevance, Classifier chains, Label Powerset). Datasets that combine heterogeneous targets (for example, binary, multi-class, and real-valued simultaneously) are not suitable for our architecture as the use of a single loss function limits us to multi-task learning problems with homogeneous targets.

Structured output and multi-class prediction problems are settings that many may consider in the multi-target prediction framework. Multi-class problems could be included using the 1-versus-rest decomposition reduction technique. Following this approach, predicting an instance's output boils down to a set of binary prediction tasks even though we remain interested in a single prediction, not multiple ones. 
Similarly to Waegeman et al.\cite{waegeman2019multi} we argue that for structured output prediction problems, the target space is often infinitely large, and the structure of the target space needs to be exploited for computational reasons during training and inference. Our framework cannot be used for structured output prediction problems where the target space cannot be enumerated (because every potential output will represent a column in the matrix representation used). As a result, we do not recommend to use our framework for problems of that kind.

Despite the extent of this work, there are still many directions we intend to explore in the future. The immediate next step will be to automate the process of hyperparameter tuning in our model. Our architecture displays unusual characteristics, like branches that can have different dimensionalities and types of sub-architectures. For that reason, the process of finding the optimal architecture using a simple technique like grid search or random search becomes practically infeasible.         
Another direction we could explore is related to the performance difference that is expected when validating in the four settings we discussed in~Section~\ref{val_setting_sec}.

Furthermore, even though in Section~\ref{multi-branch_NN_architecture} we describe both a two-branch and a tri-branch architecture, we only report experimental results using two branches. This interesting, but at the same time quite underdeveloped area of dyadic information could be an option for our future work. The collection of MTP datasets that also contain dyadic information, combined with benchmark results produced by our DeepMTP framework, would give the necessary boost to other researchers to engage in this task.

Finally, the current version of our framework optimizes specific versions of loss functions that are cell-decomposable (Hamming loss). However, in our results section, we also compared with other methods that do not optimize of the same loss as DeepMTP. An attractive next step for our work would also be to extend the range of loss functions that our DeepMTP framework allows to optimize.

\appendix
\section{Appendix}\label{definitions_appendix}

The standard MTP problem settings include multi-label classification, multivariate regression and multi-task learning and are formally defined below. 

\begin{definition}
The \textbf{multi-label classification} setting is an instance of the MTP framework with the following additional properties:
\begin{itemize}[leftmargin=*,align=left]
\itemindent=10pt    \item [(P4)] All targets are observed during training ($|\mathcal{T}| = m$).
\item [(P5)] No side information is available for targets, thus we identify them with natural numbers ($\mathbf{t}_j = j$).
 \item [(P6)] The score matrix $\mathbf{Y}$ is fully observed.
 \item [(P7)] The score set is $\mathcal{Y}=\{ 0, 1 \}$.
\end{itemize}
\end{definition}

\begin{definition}
The \textbf{multivariate regression} setting is an instance of the MTP framework with the following additional properties:
\begin{itemize}[leftmargin=*,align=left]
\itemindent=10pt    \item [(P4)] All targets are observed during training ($|\mathcal{T}| = m$).
   \item [(P5)] No side information is available for targets, thus we identify them with natural numbers ($\mathbf{t}_j = j$).
   \item [(P6)] The score matrix $\mathbf{Y}$ is fully observed.
   \item [(P7a)] The score set is $\mathcal{Y} = \mathbb{R}$.
\end{itemize}
\end{definition}

\begin{definition}
The \textbf{multi-task learning} setting is an instance of the MTP framework with the following additional properties:
\begin{itemize}[leftmargin=*,align=left]
\itemindent=10pt    \item [(P4)] All targets are observed during training ($|\mathcal{T}| = m$).
   \item [(P5)] No side information is available for targets, thus we identify them with natural numbers ($\mathbf{t}_j = j$).
   \item [(P6a)] The score matrix $\mathbf{Y}$ has missing values.
    \item [(P7a)] The score set is homogeneous across the columns of $\mathbf{Y}$.
\end{itemize}
\end{definition}

From the properties of the three definitions presented above, we see that multi-task learning can be interpreted as a generalization of multi-label classification and multivariate regression, the main difference being the missing values in the score matrix $\mathbf{Y}$. A common characteristic of the three settings presented above is that they do not use side information for the targets. The utilization of such information leads to the establishment of extensions for these three standard settings with new titles and separate research areas.

The following definitions correspond to MTP problem settings that are able to utilize side information about the target space.

\begin{definition}
The \textbf{hierarchical multi-label classification} setting is an instance of the MTP framework that shares all the properties of multi-label classification with the following updates:
\begin{itemize}[leftmargin=*,align=left]
\itemindent=10pt   \item [(P5*)] Side information is available for the targets ($\mathcal{T} = \{ \mathbf{t}_1,\ldots,\mathbf{t}_m \}$) in the form of target relations (usually hierarchies).
\end{itemize}
\end{definition}

\begin{definition}
The \textbf{dyadic prediction} setting is an instance of the MTP framework that shares all the properties of multi-task learning with the following updates:
\begin{itemize}[leftmargin=*,align=left]
\itemindent=10pt  \item [(P5*)] Side information is available for the targets ($\mathcal{T} = \{ \mathbf{t}_1,\ldots,\mathbf{t}_m \}$) in the form of a structured representation.
\end{itemize}
\end{definition}

\begin{definition}
The \textbf{zero-shot learning} setting is an instance of the MTP framework that shares all the properties of dyadic prediction setting with the following updates:
\begin{itemize}[leftmargin=*,align=left]
\itemindent=10pt    \item [(P4*)] Novel targets are expected at prediction time ($|\mathcal{T}|= m^* >m$).
\end{itemize}
\end{definition}

Properties P5 and P6 introduce the notion of inductiveness and transductiveness for instances and targets. For example, the three standard MTP problem settings are inductive w.r.t.~instances and transductive w.r.t.~targets, as predictions need to be produced for novel instances but not for novel targets. The utilization of side information for the targets is what gives the extended MTP problem settings the ability to generate predictions for novel targets. The following definitions showcase MTP problem settings that arise from the availability of side information for both instances and targets, as well as the intent to generalize to novel instances and/or targets

\begin{definition}
The \textbf{matrix completion} setting is an instance of the MTP framework that shares properties P4, P5 and P6a with the standard MTP problem settings but also has the following additional properties:
\begin{itemize}[leftmargin=*,align=left]
\itemindent=10pt     
\item [(P8)] All instances are observed during training ($|\mathcal{X}| = n$).
\item [(P9)] No side information is available for instances, thus we identify them with natural numbers ($\mathcal{X} = \{ 1,...,n \}$).
\end{itemize}
\end{definition}

In the matrix completion setting side information for instances and targets is missing, so the only achievable task is to complete the missing scores between instances and targets that are already observed during training. In these cases, matrix factorization methods utilize the structure of the score matrix $\mathbf{Y}$ in order to make predictions.

\begin{definition}
The \textbf{hybrid matrix completion} setting is an instance of the MTP framework that updates all the properties of the standard matrix completion setting:
\begin{itemize}[leftmargin=*,align=left]
\itemindent=10pt   \item [(P4*)] Novel targets are expected at prediction time ($|\mathcal{T}|= m^* >m$).
    \item [(P5*)] Side information is available for the targets ($\mathcal{T} = \{ \mathbf{t}_1,...,\mathbf{t}_m \}$) in the form of structured, hierarchical or feature representations.
   
    \item [(P8*)] Novel instances are expected at prediction time ($|\mathcal{X}|= n^* >n$).
   \item [(P9*)] Side information is available for the targets ($\mathcal{X} = \{ \mathbf{x}_1,...,\mathbf{x}_n \}$) in the form of structured, hierarchical or feature representations.
\end{itemize}
\end{definition}

Hybrid matrix completion extends the standard matrix completion setting by generalizing to novel instances and targets using the structure of the score matrix as well as side information. Both versions of the matrix completion method have been considered with great success in areas such as recommender systems, social network analysis and biological network inference. This wide adoption has also resulted in the establishment of new terms like collaborative filtering and link prediction.

\section{Experimental setup \& hyper-parameters}\label{hyperparameters_appendix}

In the following section, we detail the methods we are comparing with, the datasets we use in a per MTP problem setting manner, and the hyperparameter space we explore. Similar to other papers in the area of deep learning, we decided to complete all the experiments using a train-test (75\%-25\%) split. In papers predating the deep learning era, 5-fold or 10-fold cross-validation is usually the standard, but this has become less popular recently for problems that involve feature learning. Especially in many of the older multi-label classification papers, which often analyze low-dimensional datasets, cross-validation is often used. The choice for train-test splitting led us to rerun all the experiments for the methods we are comparing with. As a consequence we could only compare to methods that have an implementation available online.

For some of the methods included in our comparison, we performed hyperparameter optimization through a grid search, and for others, we used the default values included in their implementation. In the cases where such optimization was performed, we created an internal validation set to facilitate the process. The same strategy was used to perform early stopping in the training process of the DeepMTP framework, so that we eliminate any chance of information leaking from the original test set.

\subsection{Multi-label classification}
\begin{itemize}
    \item Multilabel k Nearest Neighbours: 
    \begin{itemize}
        \item Number of neighbours of each input instance to take into account (k): [1-10]
        \item Smoothing parameter (s): [0.5, 0.7, 1]
    \end{itemize}
    
    \item Binary Relevance (support vector machine):
    \begin{itemize}
        \item Regularization parameter (C): [0.01, 0.1, 1, 10, 100]
        \item Kernel: [linear, rbf]
    \end{itemize}
    
    \item Binary Relevance (Multilayer Perceptron):
    \begin{itemize}
        \item Learning rate: [0.001, 0.01, 0.1]
        \item Solver: [stochastic gradient decent, adam]
        \item hidden sizes: [(32), (64), (128), (256), (512)]
    \end{itemize}
    
    \item Multilayer Perceptron:
    \begin{itemize}
        \item Learning rate: [0.001, 0.01, 0.1]
        \item Solver: [stochastic gradient decent, adam]
        \item hidden sizes: [(32), (64), (128), (256), (512), (32,32), (64,64), (128,128), (256,256), (512,512)]
    \end{itemize}
    
    \item Multioutput Decision Tree Classifier:
    \begin{itemize}
        \item Split criterion: [gini, random]
        \item Splitting strategy: [best, random]
        \item The minimum number of samples required to be at a leaf node: [1, 2, 3, 4, 5]
    \end{itemize}
    
    \item Ensemble of Classifier Chains (ECC(support vector machine)):
    \begin{itemize}
        \item size of ensemble: 50
        \item Regularization parameter (C): [0.01, 0.1, 1, 10, 100]
        \item Kernel: [linear, rbf]
    \end{itemize}
    
\end{itemize}

\subsection{Multivariate regression}
\begin{itemize}
    \item Support Vector Regressor/target:
    \begin{itemize}
        \item Regularization parameter (C): [0.01, 0.1, 1, 10, 100]
    \end{itemize}
    
    \item Kernel Ridge Regressor/target:
    \begin{itemize}
        \item Regularization strength (a): [0.01, 0.1, 1, 10, 100]
        \item Gamma parameter: [0.01, 0.1, 1, 10, 100]
        \item kernel: rbf
    \end{itemize}
    
    \item Multilayer Perceptron:
    \begin{itemize}
        \item Learning rate: [0.001, 0.01, 0.1]
        \item Solver: [stochastic gradient decent, adam]
        \item hidden sizes: [(32), (64), (128), (256), (512), (32,32), (64,64), (128,128), (256,256), (512,512)]
    \end{itemize}
    
    \item Multioutput Decision Tree Regressor:
    \begin{itemize}
        \item Split criterion: [mse, mae]
        \item Splitting strategy: [best, random]
        \item The minimum number of samples required to be at a leaf node: [1, 2, 3, 4, 5]
    \end{itemize}
    
    \item Ensemble of Regressor Chains (ERC(support vector regressor)):
    \begin{itemize}
        \item Regularization parameter (C): [0.01, 0.1, 1, 10, 100]
    \end{itemize}
\end{itemize}

\subsection{Hierarchical multi-label classification}
\begin{itemize}
    \item ML-GCN: used the provided default parameters
\end{itemize}

\subsection{Matrix completion}
Used the default parameters provided in~\cite{microsoft_recommenders} for the following methods:

\begin{itemize}
    \item Matrix factorization approach by alternating least squares (MF-ALS)
    \item Fastai
    \item Riemannian  Low-rank  Matrix  Completion  (rlrmc)
    \item Wide \& deep
\end{itemize}

\subsection{Multi-task learning}
\begin{itemize}
    \item majority voting: no parameters to tune
    \item single-target Resnet: used the default parameters
\end{itemize}

\subsection{Dyadic prediction}
\begin{itemize}
    \item eBICT: 
    \begin{itemize}
        \item number of trees: 200
    \end{itemize}
\end{itemize}

\section*{Declarations}

\begin{itemize}[leftmargin=*,align=left]

\item[\textbf{Funding}]   This research received funding from the Flemish Government under the “Onderzoeksprogramma Artifici\"ele Intelligentie (AI) Vlaanderen” programme.

\item[\textbf{Conflicts of interest}]  The authors declare that they have no confict of interest.

\item[\textbf{Ethics approval}]  Not applicable.

\item[\textbf{Consent to participate}]  Not applicable.

\item[\textbf{Consent for publication}]  Not applicable.

\item[\textbf{Availability of data and material}]  The data used for the experiments are available online, see \hyperref[sec:sec6]{Section 6} for more details.

\item[\textbf{Code availability}] The code used to run the experiments can be found on github\footnotemark. The implementation of deepMTP will also be uploaded to the same repository in the near future.

\item[\textbf{Authors' contributions}]  All authors contributed equally to this work.
\end{itemize}

\footnotetext{\url{https://github.ugent.be/diliadis/deepMTP\_comparisons}}

\bibliographystyle{spmpsci}
\bibliography{bibliography}

\end{document}